\pdfoutput=1

\documentclass[11pt]{article}

\usepackage[final]{acl}
\usepackage{multirow}
\usepackage{times}
\usepackage{latexsym}
\usepackage{booktabs}
\usepackage[table, dvipsnames]{xcolor}	
\usepackage{tcolorbox}
\usepackage{arydshln}
\usepackage{amssymb}
\usepackage[T1]{fontenc}
\usepackage{float}
\usepackage{CJKutf8}
\usepackage{makecell}

\usepackage[utf8]{inputenc}

\usepackage{microtype}

\usepackage{inconsolata}

\usepackage{graphicx}

\definecolor{yellowback}{RGB}{255,243,198}
\definecolor{blueback}{RGB}{215,232,255}
\definecolor{redback}{RGB}{255,204,203}
\definecolor{greenback}{RGB}{213,231,208}

\title{SoT: Structured-of-Thought Prompting Guides Multilingual Reasoning in Large Language Models}

\author{
 \textbf{Rui Qi\textsuperscript{1,2}},
 \textbf{Zhibo Man\textsuperscript{1,2}},
 \textbf{Yufeng Chen\textsuperscript{1,2}},
 \textbf{Fengran Mo\textsuperscript{3}},
 \textbf{Jinan Xu\textsuperscript{1,2}},
\textbf{Kaiyu Huang\textsuperscript{1,2}\footnotemark[2]}
\\
 \textsuperscript{1}Key Laboratory of Big Data \& Artificial Intelligence in Transportation \\
 (Beijing Jiaotong University), Ministry of Education \\
 \textsuperscript{2}School of Computer Science and Technology, Beijing Jiaotong University \\
 \textsuperscript{3}University of Montreal\\
 \texttt{\{cherry,kyhuang\}@bjtu.edu.cn}
}

\begin{document}
\begin{CJK*}{UTF8}{gkai}
\maketitle
\renewcommand{\thefootnote}{\fnsymbol{footnote}}
\footnotetext[2]{Kaiyu Huang is the corresponding author.}

\renewcommand{\thefootnote}{\arabic{footnote}}
\setcounter{footnote}{0}
\begin{abstract}
Recent developments have enabled Large Language Models (LLMs) to engage in complex reasoning tasks through deep thinking.
However, the capacity of reasoning has not been successfully transferred to non-high-resource languages due to resource constraints, which struggles with multilingual reasoning tasks. 
To this end, we propose Structured-of-Thought (SoT), a training-free method that improves the performance on multilingual reasoning through a multi-step transformation: Language Thinking Transformation and Structured Knowledge Transformation. 
The SoT method converts language-specific semantic information into language-agnostic structured representations, enabling the models to understand the query in different languages more sophisticated. 
Besides, SoT effectively guides LLMs toward more concentrated reasoning to maintain consistent underlying reasoning pathways when handling cross-lingual variations in expression.
Experimental results demonstrate that SoT outperforms several strong baselines on multiple multilingual reasoning benchmarks when adapting to various backbones of LLMs. It can also be integrated with other training-free strategies for further improvements. 
Our code is available at \url{https://github.com/Cherry-qwq/SoT}.
\end{abstract}

\section{Introduction}
\begin{figure}[!t]
    \centering
    \includegraphics[width=1\linewidth]{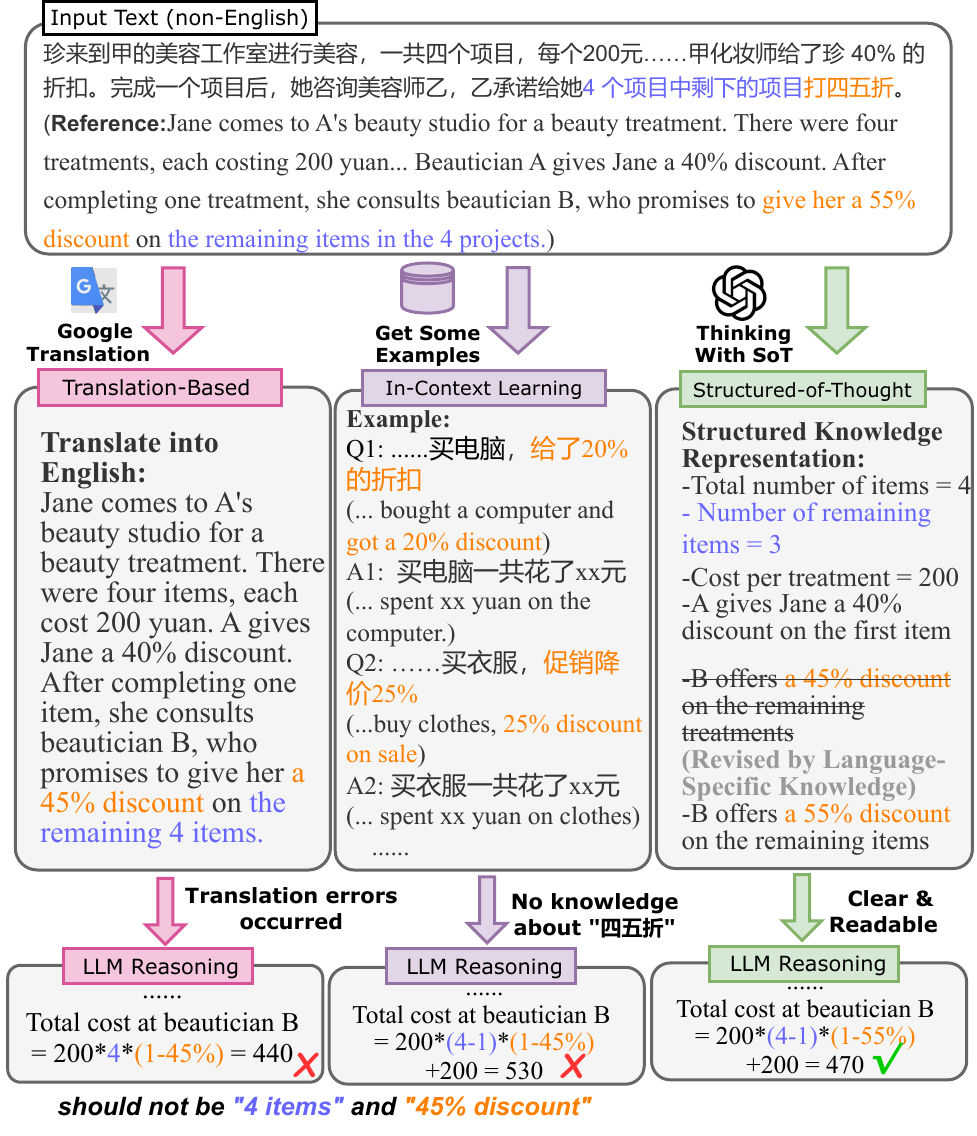}
    \caption{Examples of multilingual mathematical reasoning. When dealing with questions with \textcolor[rgb]{0.4, 0.4, 1}{complex semantic structures} and \textcolor[rgb]{1, 0.5, 0}{language-specific expressions}, LLM generate correct and incorrect answers using different prompts in non-English languages.}
    \label{fig:examples}
\end{figure}
Large language models~(LLMs) have demonstrated exceptional performance in a wide range of tasks~\citep{radford2019language, huang2025surveylargelanguagemodels}, especially in enhancing reasoning abilities~\citep{ICL}. 
Although the existing LLMs demonstrate multilingual understanding ability,
a performance gap 
is observed between different languages. This is because most large-scale datasets used for model training are predominantly available in widely spoken languages, such as English and Mandarin~\citep{huang-etal-2023-languages,shi2023language}.

An intuitive solution to mitigate this gap is to supplement multilingual data for post-training~\citep{huang2025surveylargelanguagemodels,zhang2025entropy}.
However, this is infeasible as it requires language-specific training corpora for each language, while many languages are inherently low-resource~\citep{ghosh2025multilingualmindsurvey, DBLP:journals/corr/abs-2501-02497,man-etal-2024-icl}.
Moreover, for each LLM, the post-training process demands substantial time and computational resources, which results in poor scalability for deployment in practice~\citep{zhu2023extrapolatinglargelanguagemodels,li-etal-2024-improving-context,mo2025uniconv,man2023wdsrl}.
Thus, a more appropriate approach is \textit{to enable LLM to enhance multilingual reasoning performance under training-free conditions}, and has drawn much attention in recent studies~\citep{li-etal-2023-making, SL-D, koo2025extracting,huang2025boosting,zhang2025lessbetterefficientmultilingual}.

In this scenario, previous methods aim to improve the multilingual understanding of LLMs by reformulating non-English queries, including translation-based strategies~\citep{huang-etal-2023-languages, shi2023language} and in-context learning~\citep{ICL, zhang-etal-2023-multilingual, asai2023buffetbenchmarkinglargelanguage, ahuja-etal-2023-mega,zhu2024multilingualmachinetranslationlarge}. 
The former approach relies on the availability of high-quality translations~\citep{bawden-yvon-2023-investigating}, whereas the latter would not be able to capture critical information and features without the provided well-crafted context~\citep{zhang-etal-2024-impact}.
An example of LLMs answering a mathematical problem with different prompts in a non-English language is shown in Figure~\ref{fig:examples}.
The complex semantic structures in non-English languages lead to misinterpretations of inter-entity relations, hindering accurate recognition of problems and consequently resulting in poor reasoning performance.
No matter how an identical mathematical question is formulated, its underlying reasoning process should be kept the same~\citep{hu-etal-2025-large-language,zhang2025entropy}.
Therefore, enabling LLMs to interpret problem statements accurately is crucial to establishing correct reasoning pathways in multilingual settings.

Considering the inherent reasoning capabilities of LLMs and the varying levels of difficulty in query comprehension, in this paper, we propose structured-of-thought~(SoT), a thinking strategy that incorporates structured representations into the reasoning pathway to mitigate the misinterpretation of 
LLMs in multilingual scenarios.
In particular, SoT elicits LLMs to align their reasoning pathways for non-English inputs with those thought in English via a multi-step transformation: Language Thinking Transformation and Structured Knowledge Transformation.
Beyond the mere conversion of language thinking, natural language queries are also converted into structured knowledge representations, allowing the LLMs to not only understand the context from the surface-level linguistic, but also can identify the underlying relational semantics, i.e., to achieve the equivalence of semantic understanding between expressions ``\textit{\underline{0.75 cakes per guest}}'' and ``\textit{\underline{1/4 of the guests will not attend}}'' in the example in Figure 2.
Besides, structured knowledge transformation can guide LLMs toward more concentrated reasoning by eliminating extraneous information that otherwise disrupts the inference process.
Experiments show that our SoT outperforms several state-of-the-art baselines on mathematical and commonsense reasoning tasks, and is applicable to a variety of backbone LLMs. 

Our contributions are summarized as: (1) We propose a Structured-of-Thought prompting method to guide LLMs to align the reasoning pathways for non-English queries, thereby enhancing the reasoning capabilities in the multilingual scenarios. (2) Our strategy can be integrated with other training-free prompting strategies, such as In-Context Learning~(ICL) and Chain-of-Thought~(CoT), which achieves the further improvement for multilingual reasoning. (3) Experiments demonstrate that our method can accurately understand the structural knowledge in queries to adapt various series of LLMs of different sizes on several multilingual reasoning benchmarks.

\begin{figure*}[!t]
    \centering
    \includegraphics[width=1\linewidth]{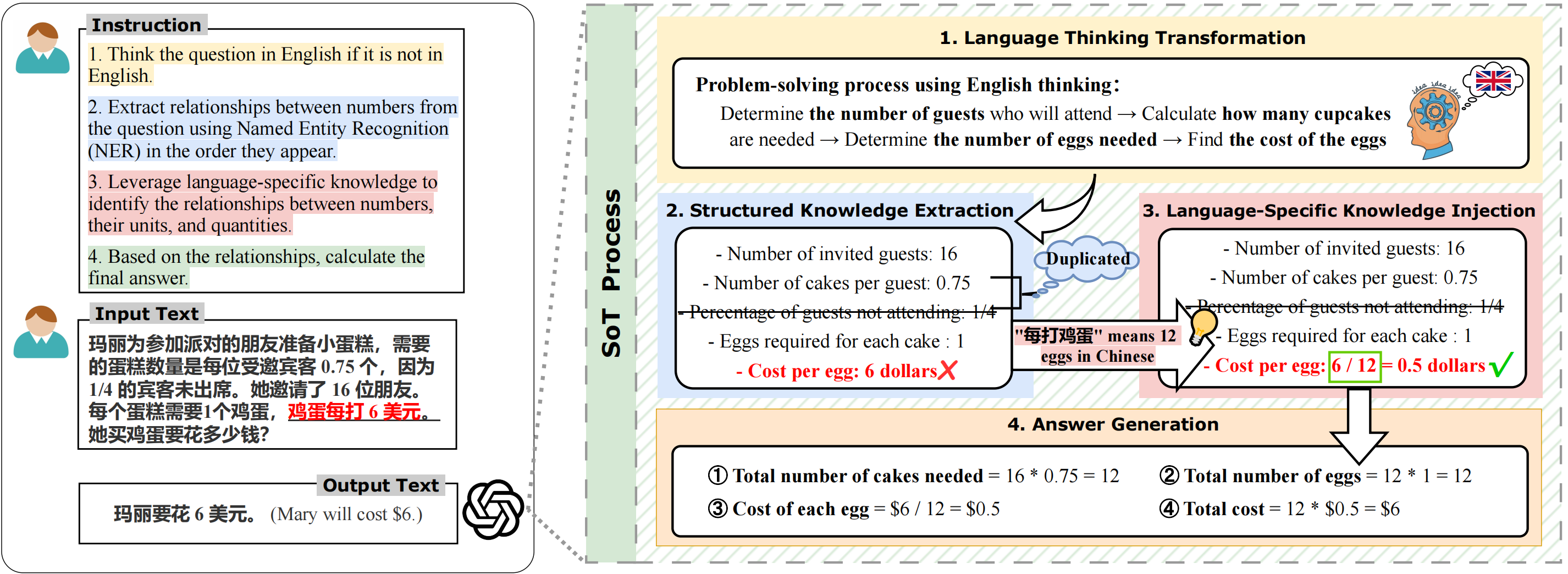}
    \caption{Overview of of the SoT strategy. The left part is an example of the question and our instructions. The right part is the thinking process of LLM under the guidance of SoT.}
    \label{fig:sot}
\end{figure*}

\section{Related Work}


\paragraph{Multilingual Reasoning.}

A common practice to enhance the multilingual reasoning capabilities of LLMs is based on 
supervised fine-tuning (SFT)~\cite{she-etal-2024-mapo,zhu-etal-2024-question,chai2025xcot}. 
However, SFT suffers from data scarcity and catastrophic forgetting, and lacks the generalization ability~\cite{she-etal-2024-mapo}.
Another research line explored the usage of carefully designed prompts to support reasoning in LLMs~\cite{huang-etal-2023-languages, qin-etal-2023-cross}.
For instance, the pre-translation approach translates input questions into a high-resource pivot language (\textit{e.g.}, English) before querying the LLM, aiming to leverage the stronger proficiency of models in the pivot language~\cite{etxaniz2024multilingual,huang2025surveylargelanguagemodels,mo2024survey}.
Furthermore, the pre-translation method can be integrated with other prompting strategies~\cite{lu2024dictionary,koo2025extracting,SL-D}, such as CoT~\cite{NEURIPS2022_9d560961} and ICL~\cite{ICL} paradigms.
Besides, \citet{liu2024translationneedstudysolving} propose several strategies to extend CoT to multilingual contexts.
Different from them, our method introduces a structured-based strategy that leverages the built-in capabilities of LLMs to mitigate the misinterpretation of semantic for multilingual reasoning.

\paragraph{Chain-of-Thought.}
CoT prompting~\cite{NEURIPS2022_9d560961,kojima2022large} is an effective step-by-step strategy for LLMs' zero-shot and few-shot reasoning.
A series of CoT-based techniques has been proposed to further improve the reasoning performance of LLMs, including Complex CoT~\cite{fu2023complexitybased,zhang2025ratt}, Decomposed Prompting~\cite{khot2022decomposed}, Multilingual CoT~\cite{shi2023language}, Least-to-Most Prompting~\cite{zhou2022least}, and Progressive-Hint Prompting~\cite{zheng2023progressive}. 
Except for exploring a CoT variant, some approaches introduce a structured representation to capture dependencies among entities for complex reasoning tasks in the thinking step~\cite{wang2024know,zhang2024blind,zhang2025ratt}.
In particular, \citet{cheng-etal-2024-structure} investigate the effectiveness of graph structure of the text in multi-step reasoning. 
Due to the limitations on foundational abilities of multilingualism~\cite{huang2025surveylargelanguagemodels}, our method attempts to exploit a more concise form to structure the knowledge in queries, which is more effective for multilingual scenarios. 

\section{Methods}


In multilingual reasoning tasks, complex semantic structures in non-English languages might obscure the relationships between entities, thereby impeding the accurate interpretation of the question. To this end, we propose SoT, a zero-shot method designed to enhance the reasoning capabilities of LLMs in multilingual scenarios through multi-step transformations. Our SoT framework consists of four steps as illustrated in Figure~\ref{fig:sot}.

The principle of our SoT is to structure the input questions by transforming reasoning pathways expressed in natural language into structured representations that are more easily interpreted by LLMs. 
This restructuring manipulation improves the abilities of models to reason accurately across languages.
In contrast to other training-free methods, SoT specifically targets the comprehension of complex semantic relationships within the questions. 
Regardless of the language in which the same question is posed, SoT allows models to fully leverage their built-in reasoning capabilities to enable LLMs to maintain correct and consistent reasoning pathways. Moreover, the framework is generalizable and can be applied across a wide range of multilingual reasoning tasks.


\subsection{Language Thinking Transformation}

\begin{tcolorbox}[colback=yellowback,  
                  colframe=black,  
                  arc=2mm,         
                  boxrule=0.5pt,   
                  width=\linewidth, 
                  boxsep=2pt, 
                  left=4pt, right=4pt,  
                  top=4pt, bottom=4pt
                  ] 
\small
\textbf{{\textit{Step 1}~}}:~~
\textit{Think the question in English if it is not in English}
\end{tcolorbox}

When the model targets the same question in different languages, its reasoning pathway should be consistent. 
Thus, we conduct the transformation of the reasoning process from low-resource to high-resource languages by cross-lingual transfer, enabling the LLMs to perform reasoning in a language in which they exhibit greater proficiency under multilingual scenarios.
In particular, we leverage the inherent reasoning and language understanding capabilities of LLMs, eliminating the need for development from scratch.
To effectively transfer the reasoning pathway into the high-resource language, we introduce a Language Thinking Transformation strategy, as illustrated in the first step in Figure~\ref{fig:sot}.

Specifically, given the sentence $X$, we 
conduct the transformation from the source language $L_s$ to the target language $L_t$~(\textit{i.e.,} English). The intermediate thinking pathways~$\mathcal{R}$ are represented as $\{r_i\}^n_{i=1}$, where $n$ denotes the number of thinking steps.
Formally, the Language Thinking Transformation process is expressed as follows:
\begin{eqnarray}
	\mathcal{R}=\arg\max p(r_1, \dots, r_n|X, L_s, L_t)
\end{eqnarray}



\subsection{Structured Knowledge Extraction}
\begin{tcolorbox}[colback=blueback,  
                  colframe=black,  
                  arc=2mm,         
                  boxrule=0.5pt,   
                  width=\linewidth, 
                  boxsep=2pt, 
                  left=4pt, right=4pt,  
                  top=4pt, bottom=4pt
                  ] 
\small
\textbf{{\textit{Step 2}~}}:~~
\textit{Extract relationships between numbers from the question using Named Entity Recognition (NER) in the order they appear.}
\end{tcolorbox}

After performing language transfer for reasoning, the knowledge from the question is extracted and then represented in a structured natural language format.
Specifically, the elements of structured knowledge mainly consist of entities and their relationship patterns.
Thus, we instruct the LLM to perform Named Entity Recognition~(NER) to identify key elements such as numerical values, units, and their associated relationships within the input text. 
The objective of this step is to construct a structured representation of knowledge~$\mathcal{K}$ that enables the LLM to accurately identify and comprehend the core entities and their interrelations within the question. 
Formally, the Structured Knowledge Extraction process is expressed as follows:
\begin{eqnarray}
	\mathcal{K}=\arg\max p(k_1, \dots, k_m|\mathcal{R},X,L_t),
\end{eqnarray}
where $\{k_i\}^m_{i=1}$ represents the pattern of structured knowledge and $m$ denotes the number of the patterns.

The construction of the structured representation eliminates irrelevant information from the input, 
making the relation among the values and entities much clearer and thus facilitating the subsequent reasoning steps with less noise.
For example, NER can facilitate the relation identification between numbers and entities in mathematical problems~(\textit{e.g.}, as shown in second step in Figure~\ref{fig:sot}, \textit{\underline{0.75 per guest}}and \textit{\underline{1/4 of guests will not attend}} represent the same relationship in different expressions).
Moreover, knowledge extraction can simplify complex problems, making them more interpretable and enhancing the capacity of LLMs to perform reasoning tasks.



\subsection{Language-Specific Knowledge Injection}

\begin{tcolorbox}[colback=redback,  
                  colframe=black,  
                  arc=2mm,         
                  boxrule=0.5pt,   
                  width=\linewidth, 
                  boxsep=2pt, 
                  left=4pt, right=4pt,  
                  top=4pt, bottom=4pt
                  ] 
\small
\textbf{{\textit{Step 3}~}}:~~
\textit{Leverage language-specific knowledge to identify the relationships between numbers, their units, and quantities.}
\end{tcolorbox}

Although the language transfer in thinking helps the LLM better interpret the problem, it neglects language-specific differences of expression in terms of 
quantities, units, and their relations. 
To address this, the third step in our approach aims to further enhance the understanding of non-English languages by guiding LLMs to focus on language-specific knowledge. 
Each language possesses unique rules and conventions for expressing numerical relations and quantities. 
For example, in Chinese, the phrase ``四五折'' denotes a 55\% discount, which might lead to misinterpretation if processed without cultural or contextual awareness.
LLMs might not be able to distinguish that they have the same meaning when performing calculations directly. An alternative is to leverage translation-based strategies as intermediate support, which would still fail to capture these nuances accurately.
Guided by language-specific expressions, the LLM can accurately understand these nuances, reducing misunderstandings caused by linguistic variation and improving reasoning performance across languages.
Formally, the Language-Specific Knowledge process is expressed as follows:
\begin{eqnarray}
	\mathcal{K}^{L_s}=\arg\max p(k^{L_s}_1, \dots, k^{L_s}_m|\mathcal{K},L_s),
\end{eqnarray}
where $\{k^{L_s}_i\}^m_{i=1}$ represents the language-specific knowledge.

\subsection{Answer Generation}
\begin{tcolorbox}[colback=greenback,  
                  colframe=black,  
                  arc=2mm,         
                  boxrule=0.5pt,   
                  width=\linewidth, 
                  boxsep=2pt, 
                  left=4pt, right=4pt,  
                  top=4pt, bottom=4pt
                  ] 
\small
\textbf{{\textit{Step 4}~}}:~~
\textit{Based on the relationships, calculate the final answer in the Source Language.}
\end{tcolorbox}

The final stage is to integrate the above information, where the LLM conducts reasoning based on the extracted structured knowledge, language-specific knowledge, and the results of the language thinking transformation, towards the final answer $\mathcal{F}$. 
The answer is transformed back into the source language~$L_s$, ensuring consistency between input and output to maintain interpretability in multilingual scenarios. Formally, the generation of the final answer is determined as:
\begin{eqnarray}
	\mathcal{F}=\arg\max p(f|\mathcal{R},\mathcal{K},\mathcal{K}^{L_s},L_s)
\end{eqnarray}




\begin{table*}[!t]
    \centering
    \small
    \begin{tabular}{lccccccccccc}
        \toprule
        \multirow{2}{*}{\textbf{Methods}}   & \multicolumn{11}{c}{\textbf{Language}} \\
        \cmidrule(lr){2-11}
         \textbf{ } & \textbf{En} & \textbf{Sw} & \textbf{Ja} & \textbf{Be} & \textbf{Th} & \textbf{Ru} & \textbf{Zh} & \textbf{De} & \textbf{Es} & \textbf{Fr} & \textbf{Avg.} \\
        \midrule
        (\textbf{\texttt{training-free}}) & \multicolumn{11}{c}{(\texttt{DeepSeek-R1-7B})}\\
        Direct & 82.0 & 18.6 & 67.8 & 52.6 & 53.8 & 80.2 & 78.0 & 73.0 & 80.4 & 71.8 & 65.8 \\
        DoLa & 83.8 & 18.7 & 70.1 & 54.0 & 62.2 & 83.0 & 81.3 & 75.3 & 80.9 & 74.0 & 68.3 \\
        SL-D & 84.1 & 22.6 & 73.1 & 55.7 & 64.2 & \underline{84.8} & 84.3 & 79.0 & 81.6 & 77.1 & 70.7 \\
        DIP & 88.0 & 21.4 & \underline{82.0} & \underline{63.5} & 64.5 & 83.2 & 85.0 & 82.1 & 83.0 & 83.4 & 73.6 \\
        CLP & \underline{89.6} & \underline{23.2} & 77.0 & 62.7 & \underline{69.3} & 78.5 & 84.8 & 81.4 & 81.8 & \underline{87.0} & 73.5 \\
        EMCEI & 89.0 & 23.0 & 80.0 & 61.0 & 64.9 & 83.8 & \underline{86.2} & \underline{83.2} & \underline{83.4} & 84.9 & \underline{73.9} \\
        \hdashline
        SoT~(\textit{Ours}) & \textbf{89.8} & \textbf{24.8} & \textbf{82.8} & \textbf{64.6} & \textbf{71.8} & \textbf{85.4} & \textbf{87.2} & \textbf{85.4} & \textbf{85.2} & \textbf{88.2} & \textbf{76.5}\\
        \midrule
        (\textbf{\texttt{post-training}}) \\
        xCoT & 84.7 & 50.7 & 79.6 & 59.0 & 64.6 & 80.3 & 83.2 & 82.7 & 85.1 & 88.3 & 75.8 \\
        QAlign & 82.8 & 46.2 & 82.6 & 56.0 & 64.5 & 81.4 & 80.3 & 86.6 & 89.8 & 89.1 & 75.9 \\
        MindMerger & 83.6 & 44.6 & 83.4 & 56.6 & 59.7 & 81.2 & 84.6 & 87.4 & 89.1 & 92.2 & 76.2 \\
        MAPO & 84.8 & 50.2 & 83.8 & 53.6 & 64.9 & 80.5 & 84.8 & 83.2 & 88.2 & 85.2 & 75.9 \\
        \midrule
        \midrule
        (\textbf{\texttt{training-free}}) & \multicolumn{11}{c}{(\texttt{Qwen2.5-7B-Instruct})}\\
        Direct & 89.8 & 39.4 & 69.2 & 55.0 & 65.4 & 74.6 & 81.8 & 77.8 & 83.2 & 82.2 & 71.8 \\
        DoLa & 91.0 & 54.4 & 73.1 & 64.7 & 74.5 & 76.4 & 83.3 & 79.2 & 85.3 & 85.7 & 76.8 \\
        SL-D & \underline{91.5} & 56.0 & 75.1 & 66.7 & \underline{77.2} & 77.1 & 85.4 & 81.6 & 88.2 & 87.5 & 78.6 \\
        DIP & 88.3 & 52.1 & 86.3 & \textbf{77.1} & 76.1 & 84.4 & \underline{87.8} & \underline{91.2} & 88.0 & \underline{90.1} & 82.1 \\
        CLP & 90.2 & 50.3 & 80.6 & 67.4 & 74.4 & 79.0 & 82.2 & 85.1 & 83.9 & 87.0 & 78.0 \\
        EMCEI & 89.6 & \underline{58.2} & \underline{86.7} & 74.6 & 75.2 & \underline{86.0} & 87.7 & 90.6 & \underline{89.4} & 89.3 & \underline{82.7} \\
        \hdashline
        SoT~(\textit{Ours}) & \textbf{93.6} & \textbf{61.0} & \textbf{87.6} & \underline{76.4} & \textbf{83.8} & \textbf{87.4} & \textbf{89.4} & \textbf{91.6} & \textbf{91.8 }& \textbf{91.2} & \textbf{85.4} \\
        \midrule
        (\textbf{\texttt{post-training}}) \\
        xCoT & 85.0 & 60.1 & 81.0 & 62.4 & 66.3 & 84.1 & 85.2 & 88.5 & 90.3 & 89.0 & 79.2 \\
        QAlign & 80.3 & 52.1 & 83.0 & 59.6 & 64.9 & 85.6 & 81.4 & 92.5 & 93.3 & 89.9 & 78.3 \\
        MindMerger & 81.5 & 51.0 & 84.5 & 58.3 & 59.6 & 83.4 & 90.9 & 89.2 & 90.0 & 93.7 & 78.2 \\
        MAPO & 84.6 & 57.6 & 85.2 & 53.0 & 68.0 & 84.2 & 84.7 & 84.5 & 88.4 & 85.4 & 77.6 \\

    \bottomrule
    \end{tabular}%
     \caption{Results~(\%) of mathematical reasoning on MSVAMP. For all \textbf{\texttt{training-free}} methods, the \textbf{bold} text represents the highest scores, while the \underline{underline} represents the second highest scores.}
    \label{tab:results_msvamp}
\end{table*}

\section{Experiments}

\subsection{Experimental Setup}

\paragraph{Models.}  We select three series of LLMs to verify the effectiveness of SoT: \textbf{\texttt{gpt-3.5-turbo}}, \textbf{\texttt{Qwen2.5-7B-Instruct}} and \textbf{\texttt{DeepSeek-R1-7B}}, including both open-source and closed-source models, ranging from past to latest. To further demonstrate the robustness on larger models, we utilize \textbf{\texttt{Qwen2.5-32B-Instruct}} as the basic model. 

\paragraph{Benchmarks and Evaluation.} 
To ensure the reliability of the experiments, all methods are implemented on two mathematical reasoning tasks~(MGSM~\citep{shi2023language} and MSVAMP~\citep{msvamp}) and one commonsense reasoning task~(XCOPA~\cite{xcopa}).
Benchmark details are listed in Appendix~\ref{sec:data_details}.
We employ the accuracy to access the ability of the methods for all tasks~\cite{jin-etal-2024-length,chen2023breaking}.

\begin{table*}[!t]
    \centering
    \small
    \begin{tabular}{lcccccccccccc}
        \toprule
        \multirow{2}{*}{\textbf{Methods}}   & \multicolumn{11}{c}{\textbf{Language}} \\
        \cmidrule(lr){2-12}
         \textbf{ } & \textbf{En} & \textbf{Sw} & \textbf{Ja} & \textbf{Be} & \textbf{Th} & \textbf{Te} & \textbf{Ru} & \textbf{Zh} & \textbf{De} & \textbf{Es} & \textbf{Fr} & \textbf{Avg.} \\
        \midrule
        (\textbf{\texttt{training-free}}) & \multicolumn{12}{c}{(\texttt{DeepSeek-R1-7B})}\\
        Direct & 75.2 & 7.2 & 42.4 & 43.6 & 41.6 & 18.0 & 65.6 & 72.0 & 50.0 & 64.0 & 55.6 & 48.7 \\
        DoLa & 75.8 & 8.0 & 43.4 & 46.4 & 45.4 & 18.0 &60.2 & 71.2 & 59.2 & 66.2 & 53.8 &  49.8 \\
        SL-D & 77.2 & 8.4 & 57.0 & 47.0 & 46.0 & 20.0 &62.4 & 72.8 & 62.6 & 64.6 & 62.0 &  52.7 \\
        DIP & 80.0 & 6.0 & 51.2 & 48.2 & \underline{57.8} & \underline{24.2} &\underline{64.8} & \underline{75.4} & 60.6 & 67.0 & 63.0 &  54.4 \\
        CLP & \textbf{87.0} & \underline{9.0} & \underline{60.4} & \underline{50.0} & 56.8 & 19.0 &61.2 & 71.8 & \underline{65.8} & 67.4 & 65.0 &  \underline{55.8} \\
        EMCEI & 81.2 & 7.2 & 58.2 & 46.4 & 57.0 & 18.4 &64.2 & 74.6 & 63.6 & \underline{68.0} & \underline{67.2} &  55.1 \\
        \hdashline
        SoT~(\textit{Ours}) & \underline{84.4} & \textbf{10.0} & \textbf{61.2} & \textbf{51.2} & \textbf{61.2} & \textbf{28.0} & \textbf{70.0} & \textbf{76.4} & \textbf{70.0} & \textbf{71.6} & \textbf{68.0} & \textbf{59.3} \\

        \midrule
        (\textbf{\texttt{post-training}}) \\
        xCoT & 82.2 & 43.4 & 62.0 & 56.6 & 55.8 & 10.0 & 71.4 & 75.4 & 67.2 & 74.2 & 67.0 & 60.5 \\
        QAlign & 81.6 & 42.8 & 60.4 & 53.6 & 53.0 & 11.4 & 69.6 & 74.0 & 68.6 & 72.2 & 66.2 & 59.4 \\
        MindMerger & 80.0 & 41.6 & 60.8 & 54.2 & 53.8 & 12.8 & 70.2 & 75.8 & 69.8 & 73.4 & 65.2 & 59.8 \\
        MAPO & 86.2 & 42.2 & 61.6 & 53.2 & 59.4 & 12.0 & 69.6 & 78.0 & 67.4 & 71.8 & 61.6 & 60.3 \\
        \midrule
        \midrule
        (\textbf{\texttt{training-free}}) & \multicolumn{12}{c}{(\texttt{Qwen2.5-7B-Instruct})}\\
        Direct & 84.0 & 12.8 & 56.0 & 51.2 & 48.0 & 24.0 & 73.6 & \textbf{80.8} & 66.8 & 71.2 & 64.4 & 57.5 \\
        DoLa & 83.2 & 13.8 & 61.0 & 61.2 & 54.4 & 32.0 & 75.4 & 75.0 & 69.2 & 73.4 & 67.8 & 60.6 \\
        SL-D & 84.6 & 10.4 & 63.2 & 63.2 & 54.4 & \underline{34.6} & 76.2 & 76.0 & 70.6 & 74.2 & 69.0 & 61.5 \\
        DIP & 84.4 & 24.2 & 70.4 & 66.8 & 64.4 & 33.2 & 78.0 & 76.6 & 70.0 & 78.0 & \underline{74.2} & 65.5 \\
        CLP & 84.2 & 20.0 & 70.8 & 64.4 & 65.4 & 30.0 & \underline{78.8} & 76.0 & 71.0 & 77.8 & 72.0 & 64.6 \\
        EMCEI & \underline{84.8} & \underline{27.0} & \underline{71.0} & \underline{68.2} & \underline{72.0} & 31.0 & 76.6 & 75.6 & \underline{72.0} & \underline{78.4} & 73.0 & \underline{66.3} \\
        \hdashline
        SoT~(\textit{Ours}) & \textbf{85.6} & \textbf{28.0} & \textbf{71.8} & \textbf{69.6} & \textbf{74.0} & \textbf{36.4} & \textbf{80.8} & \underline{77.0} & \textbf{72.8} & \textbf{79.6} & \textbf{75.2} & \textbf{68.3} \\
        \midrule
        (\textbf{\texttt{post-training}}) \\
        xCoT & 85.6 & 47.2 & 64.2 & 62.2 & 61.8 & 12.6 & 79.4 & 85.2 & 70.2 & 79.0 & 78.2 & 66.0 \\
        QAlign & 84.6 & 45.8 & 60.8 & 61.4 & 62.4 & 13.2 & 75.8 & 81.6 & 72.0 & 72.6 & 73.2 & 63.9 \\
        MindMerger & 82.4 & 44.4 & 62.4 & 56.2 & 59.4 & 12.0 & 79.0 & 85.4 & 70.0 & 69.2 & 69.8 & 62.7 \\
        MAPO & 88.4 & 46.0 & 63.0 & 58.8 & 62.2 & 12.4 & 78.3 & 88.3 & 68.2 & 71.0 & 68.0 & 64.1\\
        \bottomrule
    \end{tabular}%
     \caption{Results~(\%) of mathematical reasoning on MGSM. For all \textbf{\texttt{training-free}} methods, the \textbf{bold} text represents the highest scores, while the \underline{underline} represents the second highest scores.}
    \label{tab:results_mgsm}
\end{table*}

\paragraph{Baselines.}  For comparison, we select recent advanced training-free methods~(\textit{e.g.}, DoLa~\cite{DOLA}, SL-D~\cite{SL-D}, DIP~\cite{lu2024dictionary}, CLP~\cite{qin-etal-2023-cross}, EMCEI~\cite{koo2025extracting}.) and effective post-training methods~(xCoT~\cite{chai2025xcot}, QAlign~\cite{zhu-etal-2024-question}, MindMerger~\cite{huangmindmerger} and MAPO~\cite{she-etal-2024-mapo}). 
We follow the original settings of the original paper.
More details of baselines are listed in Appendix~\ref{app_baselines}.


     
     

\begin{figure}[!t]
    \centering
    \includegraphics[width=0.8\linewidth]{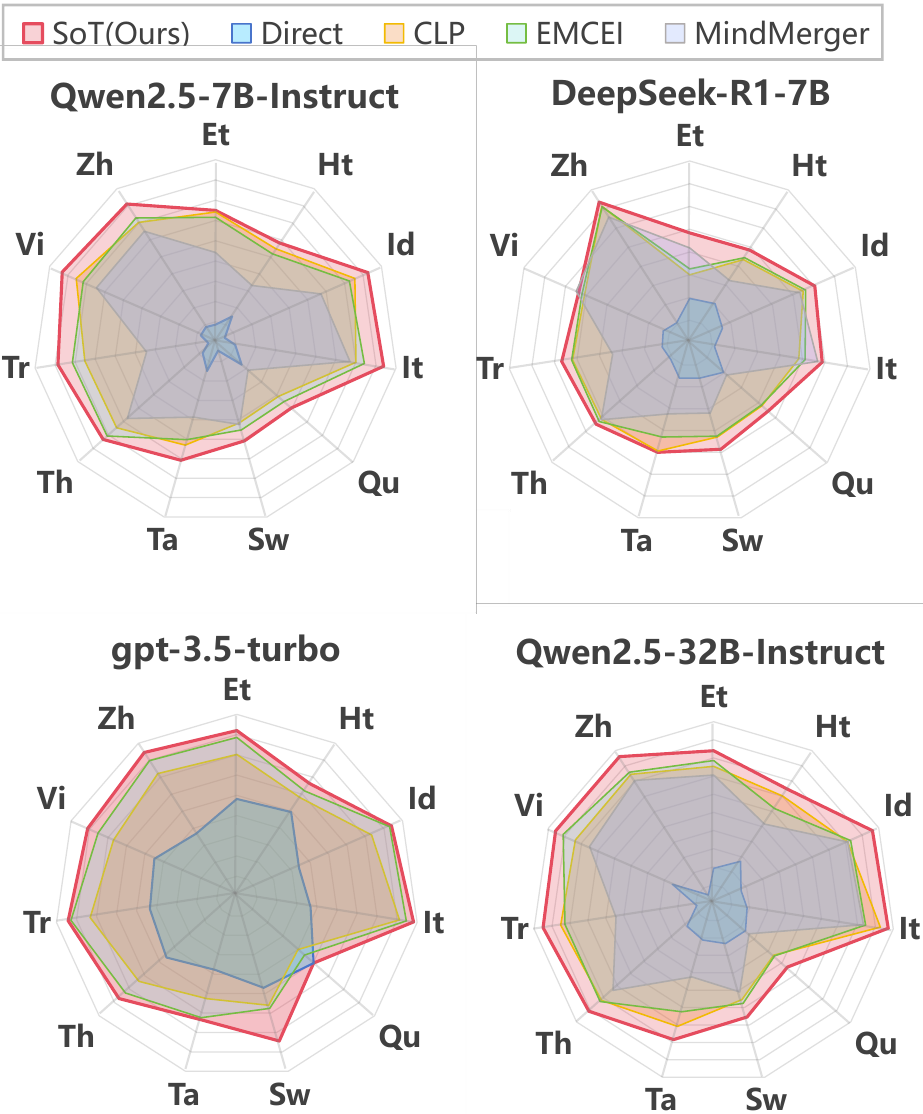}
    \caption{Results of commonsense reasoning on XCOPA using various LLMs. }
    \label{xcopa}
\end{figure}

\subsection{Main Results}


\paragraph{Performance on Mathematical Reasoning.}

As shown in Table~\ref{tab:results_msvamp} and Table~\ref{tab:results_mgsm}, 
we investigate the mathematical abilities of LLMs with different methods that facilitate multilingualism across various languages.
The results demonstrate that our proposed method (SoT) outperforms several baselines in terms of average accuracy, including the training-free and post-training methods.
The training-free methods focus on stimulating the inherent knowledge of the foundational LLMs, which can achieve gains in most languages with decreasing cost. 
However, due to the inherent defects of the model, it is difficult to achieve significant improvement for languages with insufficient inherent knowledge of the model. 
Although the post-training methods can alleviate this issue, these methods face limitations in data construction, where the effects achieved in different languages and tasks are unstable.
Moreover, the more strengthful model diminishes the effectiveness of post-training methods, which opposes the core advantages of our method. 
All the results using various LLMs are listed in Appendix~\ref{sec:all_results}.

\paragraph{Performance on Commonsense Reasoning.}
As shown in Figure ~\ref{xcopa}, we also investigate the effectiveness of SoT on the commonsense reasoning task, compared with other methods.
The results demonstrate that the advantages of SoT are further enhanced, which has an obvious improvement over the original method~(\textit{Direct}), compared with other baselines.
In particular, the structured knowledge in our method can not only extract the computational relationships for reasoning in mathematical problems, but also enable LLMs to deeply think about the logical relationships between entities in commonsense reasoning.
The post-training method does not show gains similar to those in the supervised direction for low-resource langauges~(\textit{e.g.,} a significant improvement on Sw) due to the limitation of the corpus, while the training-free methods demonstrate better generalization, especially SoT.

\begin{table}[!t]
\begin{centering}
\small
\begin{tabular}{ccccc}

\toprule
\multirow{2}{*}{No.} & \multicolumn{3}{c}{Multi-Step Scopes}  & \multirow{2}{*}{Avg.}\\
\cmidrule(r){2-4}
 & Step 1 & Step 2 & Step 3 & \\
\midrule
1 & $\times$ & $\times$ & $\times$ & 37.3  \\
2 & \checkmark & $\times$ & $\times$ & 40.0  \\
3 & $\times$ & \checkmark & $\times$ & 53.2  \\
4 & $\times$ & $\times$ & \checkmark & 58.1  \\
5 & \checkmark & \checkmark& $\times$ &  60.8 \\
6 & \checkmark & $\times$ & \checkmark & 61.2  \\
7 & $\times$ & \checkmark & \checkmark &  61.6 \\
8 & \checkmark & \checkmark & \checkmark &  \textbf{62.8} \\
\bottomrule
\end{tabular}
\par\end{centering}
\caption{Results of different prompting strategies on MGSM and \textbf{\texttt{gpt-3.5-turbo}} in terms of average scores. 
}
\label{tab:ablation}
\vspace{-2ex} 
\end{table}

\begin{figure}[!t]
    \centering
    \includegraphics[width=1\linewidth]{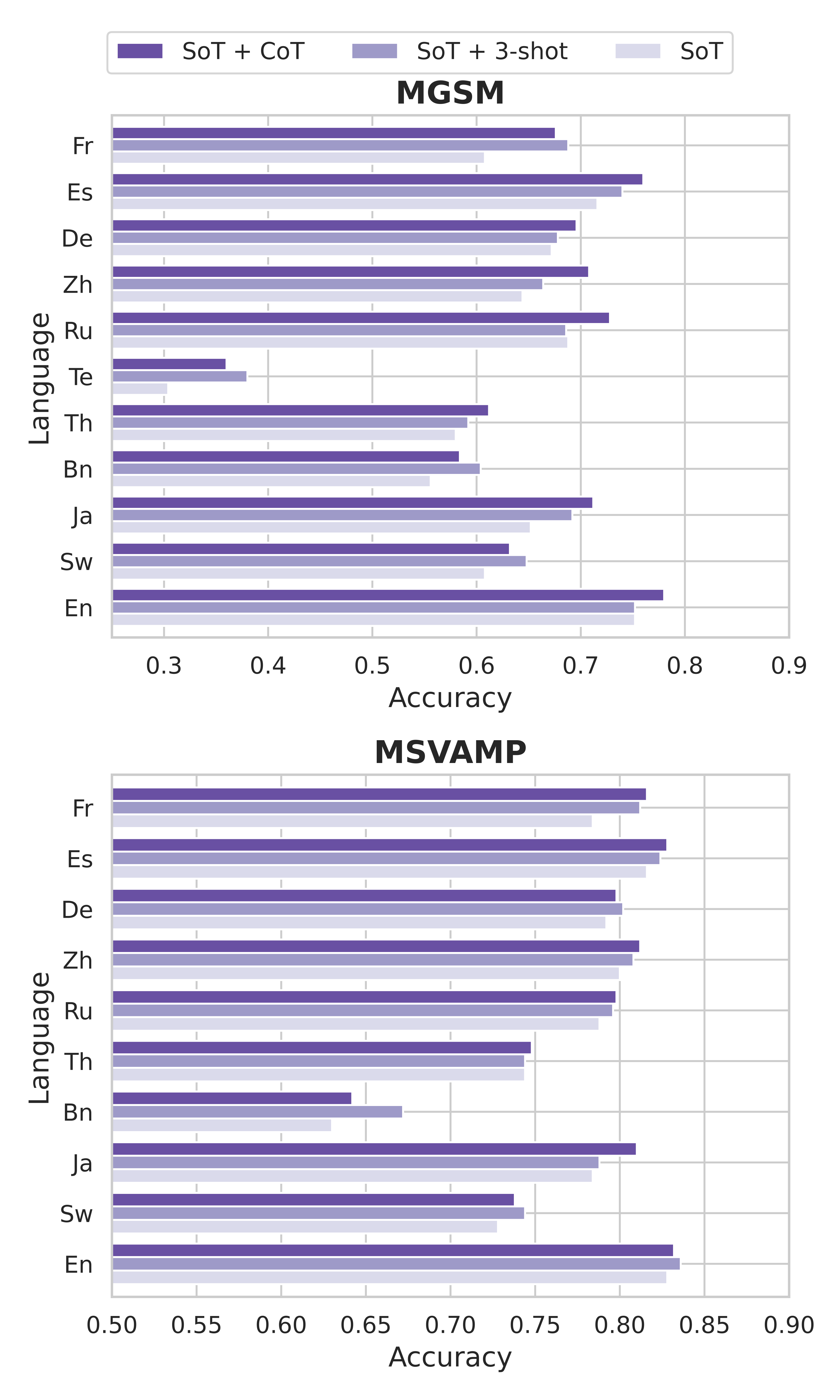}
    \caption{Results of SoT combined with CoT and few-shot~(3-shot) on the MGSM and MSVAMP using \textbf{\texttt{gpt-3.5-turbo}}.}
    \label{many_method}

\end{figure}

\subsection{Ablation Studies}

\paragraph{Effects of Muti-Step Scopes.}

As shown in Table~\ref{tab:ablation}, we explore the contribution of each step in SoT.
The results demonstrate that our method can help queries in diverse languages to be better understood and achieves better performance when both three steps are considered through SoT for mathematical reasoning in the multilingual scenarios. 
Specifically, each individual step in SoT has a positive impact, according to the comparison among the Strategies No.1, 2, 3 and 4.
Furthermore, the two combined forms further enhance the reasoning performance in terms of the Methods No.5, 6 and 7.
Except for SoT~(No.8), the results show that the structured extraction and language-specific knowledge~(No.7) are more important and achieve the highest performance~(61.6\%), indicating that language transfer thinking has a positive impact, but is not an indispensable factor.
    
\begin{figure}[!t]
    \centering
    \includegraphics[width=1\linewidth]{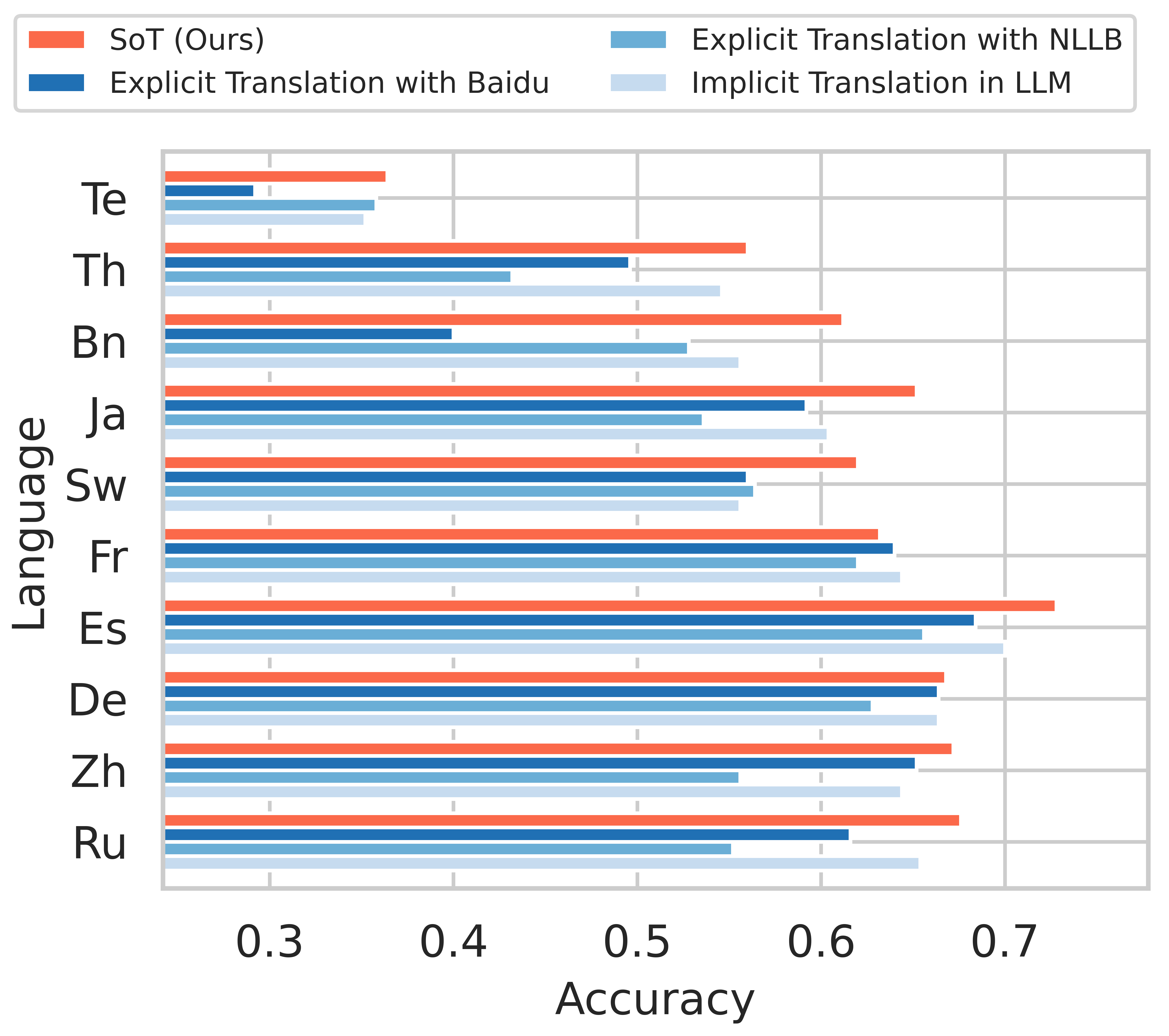}
    \caption{Results on thinking and translation. \textcolor{red}{SoT} employs the thinking manner, while other methods replace the language thinking with \textcolor{blue}{3 translation processes}.}
    \label{vs_translation}
\end{figure}


\paragraph{Effects on Boosting Multilingual Capability.}
Since SoT has shown strong performance across all languages, it's worthwhile to explore whether the improvement reflects enhanced general reasoning or a true boost in multilingual capability. Strategies No.1 and 2 in Table~\ref{tab:ablation}  also show the performance of SoT on transferred-language inputs and native-language inputs. The results show that SoT can achieve competitive performance on native-language inputs, but it is worse than adding language transformation, which is attributed to the model's sufficient training in English.
It confirms that SoT enhances multilingual capability rather than merely general reasoning ability through language transformation.

\paragraph{Effects of Integrated Methods.}
As shown in Figure ~\ref{many_method}, we explore the feasibility of SoT when combined with other training-free methods such as CoT and ICL.
The results show that the adoption of CoT or ICL further improves SoT performance, demonstrating that SoT does not have conflicts with other training-free methods.
Specifically, CoT achieves better performance in high-resource languages, while ICL is more proficient in low-resource languages.
A possible reason is that CoT is suited to guide intrinsic knowledge in LLMs and ICL provides the language knowledge which is a supplement to low-resource languages for LLMs. More comparison is shown in Appendix~\ref{sec:all_results}.

\begin{figure*}[!t]
    \centering
    \includegraphics[width=\linewidth]{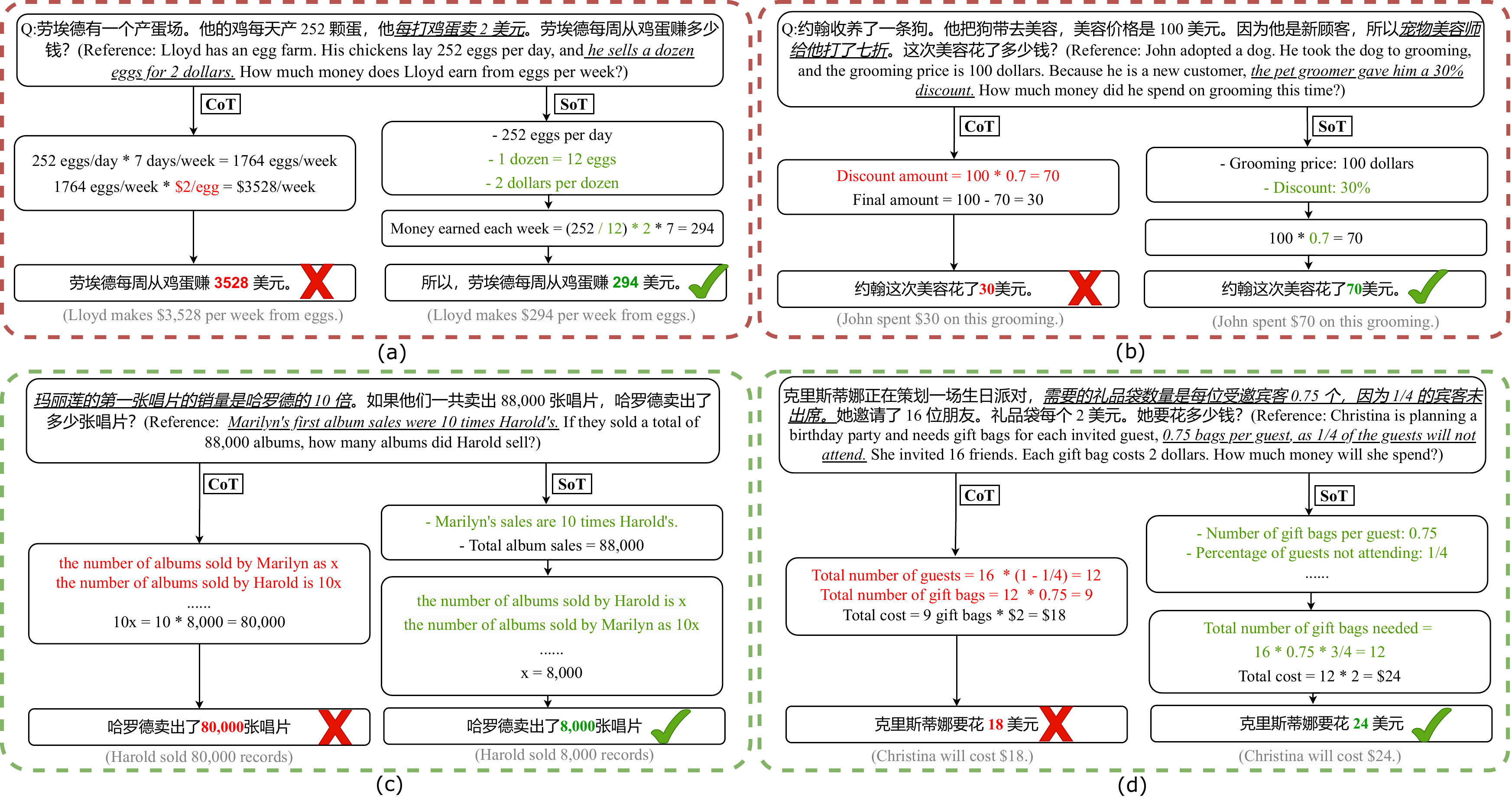}
    \caption{Examples of CoT and SoT on the mathematical reasoning tasks. We only highlight some words and fragments to show the representative difference between the two methods. The \textcolor{red}{red} parts represent the misunderstanding, while the \textcolor[rgb]{0, 0.6, 0}{green} parts represent a correct understanding.}
    \label{fig:case_study}
\end{figure*}

\subsection{Results on Thinking and Translation}
As shown in Figure~\ref{vs_translation}, we explore the effectiveness of \textit{Step 1}~(Language Thinking Transformation) which is replaced with the translation process.
Previous studies attempt to translate original queries into a high-resource language~(e.g., English), which avoids the problem of insufficient abilities in the source language.
Formally, we modify the instruction of \textit{Step 1} as follow:

\begin{tcolorbox}[colback=lightgray!10,  
                  colframe=black,  
                  arc=2mm,         
                  boxrule=0.5pt,   
                  width=\linewidth, 
                  boxsep=2pt, 
                  left=4pt, right=4pt,  
                  top=4pt, bottom=4pt
                  ] 
\small
\textbf{\textcolor{red}{\textit{Step 1}~(\textit{Thinking} $\Rightarrow$ \textit{Translation})}}:\\
\textit{Type~1: Translate the question into English if it is not in English for the following step.\\
Type~2: \colorbox [HTML]{FFD966}{[Outputs by Translators $T_s$]} is the translation of question for the following step.}
\end{tcolorbox}
We divide the translation methods into two types: The first is to replace ``thinking'' in the instruction with ``translate'' for implicit translation~(\textit{i.e.,} no translation result is generated).
The second is to replace the instruction with the explicit translation by the external translator.
The results reveal that the robustness of thinking transformation is better than that of translation, in which the reasoning performance of the translation manner is influenced by the translation qualities of the source language.
Translation errors will accumulate and be passed on to subsequent steps via either implicit translation~(LLM translator) or explicit translation~(Baidu or NLLB translator), causing performance degradation.

\begin{table}[ht]
\centering
\small
\begin{tabular}{ccccc}
\toprule
\multirow{2}{*}{\textbf{Model}} & \multirow{2}{*}{\textbf{Dataset}} & \multicolumn{3}{c}{\textbf{Decoding Time(s)}} \\
\cmidrule(lr){3-5}
 & & \textbf{Direct} & \textbf{EMCEI} & \textbf{SoT} \\
\midrule
\multirow{2}{*}{\makecell{Qwen2.5-7B-\\Instruct}} & MGSM & 3.922 & 5.134 & 4.168 \\
 & XCOPA & 3.184 & 4.570 & 3.457 \\
\midrule
\multirow{2}{*}{\makecell{Deepseek-\\R1-7B}} & MGSM & 3.697 & 4.881 & 3.919 \\
 & XCOPA & 3.216 & 4.337 & 3.544 \\
\bottomrule
\end{tabular}
\caption{Average decoding time of SoT in all languages combined with direct method and comparison method like EMCEI.}
\label{table:decoding_time}
\end{table}

\begin{table*}[ht]
\centering
\footnotesize  
\resizebox{\textwidth}{!}{
\begin{tabular}{ccccccccccc}
\toprule
\multirow{2}{*}{\textbf{Model}} & \multirow{2}{*}{\textbf{Dataset}} & \multicolumn{3}{c}{\textbf{Input Tokens}} & \multicolumn{3}{c}{\textbf{Output Tokens}} & \multicolumn{3}{c}{\textbf{Total Tokens}} \\
\cmidrule(lr){3-11}
 & & \textbf{Direct} & \textbf{EMCEI} & \textbf{SoT} & \textbf{Direct} & \textbf{EMCEI} & \textbf{SoT} & \textbf{Direct} & \textbf{EMCEI} & \textbf{SoT} \\
\midrule
\multirow{2}{*}{Qwen2.5-7B-Instruct} & MGSM & 115.7 & 255.5 & 193.3 & 924.5 & 1118.7 & 930.3 & 1040.2 & 1374.2 & 1123.6 \\
 & XCOPA & 86.7 & 213.1 & 184.7 & 771.7 & 843.8 & 630.3 & 858.4 & 1056.8 & 815.0 \\

\multirow{2}{*}{Deepseek-R1-7B} & MGSM & 116.6 & 217.6 & 194.3 & 646.8 & 946.7 & 770.7 & 763.5 & 1164.3 & 965.0 \\
 & XCOPA & 87.7 & 213.8 & 185.7 & 518.4 & 679.5 & 555.2 & 606.1 & 893.3 & 741.0 \\
\bottomrule
\end{tabular}}
\caption{Average input, output and total tokens of SoT in all languages combined with direct method and comparison method like EMCEI.}
\label{table:performance_metrics}
\end{table*}

\subsection{Analysis of Resource Consumption}

\paragraph{Time Consumption.}
As shown in Table~\ref{table:decoding_time}, we explore the gap in inference time between SoT and other methods like direct method and EMCEI.
Analysis reveals that SoT achieves a shorter decoding time compared to other effective methods like EMCEI, while introducing only a slight increase in inference time compared to the direct method. SoT adds an average of just 0.1 seconds, which is nearly negligible.
\paragraph{Token Consumption.}
As shown in Table~\ref{table:performance_metrics}, we explore the token consumption between SoT and other representative methods.
Results show that SoT achieves lower input and output token consumptions compared to EMCEI. Although SoT introduces slightly higher token consumptions than the direct method, the performance gains are more substantial.
Moreover, existing LLMs have optimization mechanisms for long context input. Thus, adding a portion of token input does not impose a burden on efficiency but can significantly enhance effectiveness.

\begin{table}[h]
\centering
\begin{tabular}{ccc}
\toprule
\multirow{2}{*}{\textbf{Dataset} } & \multicolumn{2}{c}{\textbf{Extracted Knowledge} } \\
\cmidrule(lr){2-3}
 & \textbf{Step 2} & \textbf{Step 3} \\
\midrule
MGSM & 3.78 & 1.80 \\
MSVAMP & 2.97 & 1.72 \\
XCOPA & 2.03 & 1.77 \\
Avg. & 2.93 & 1.76 \\
\bottomrule
\end{tabular}
\caption{ The number of knowledge extracted by each steps of SoT on the MGSM, MSVAMP and XCOPA using gpt-3.5-turbo.}
\label{table:step_number}
\end{table}


\subsection{Details of Extracted Knowledge and Relations}
As shown in Table~\ref{table:step_number}, a quantitative analysis is performed on the knowledge extracted in the "Structured Knowledge Extraction" step (Step 2) and "Language-Specific Knowledge Injection" step (Step 3). Experimental results show that SoT extracts about 2.93 structured knowledge instances in the step 2, and then insert about 1.76 language-specific instances in the step 3 to adjust the correction of language knowledge.
The application frequency is approximately three-fifths.

\subsection{Case Study}
As shown in  Figure~\ref{fig:case_study}, we present examples in MGSM where the traditional CoT method fails, while our framework produces accurate results.
The cases highlight the effectiveness of our approach to resolve common errors in multilingual reasoning. Examples in MSVAMP and XCOPA can be found in Appendix~\ref{sec:case_study}.

As shown in Figure~\ref{fig:case_study}.a and Figure~\ref{fig:case_study}.b, CoT suffers from misinterpreting units and discounts due to language-specific ambiguities. 
For instance, in Figure~\ref{fig:case_study}.a, CoT confuses ``per dozen eggs'' with ``per egg'', leading to an incorrect calculation. Similarly, in Figure~\ref{fig:case_study}.b, the expression ``70\% off'' is misunderstood by CoT as ``a 70\% reduction'' in Chinese, rather than ``70\% of the original price''. 
SoT effectively resolves these issues by incorporating structured and language knowledge, ensuring correct numerical interpretation.

As shown in Figure~\ref{fig:case_study}.c and Figure~\ref{fig:case_study}.d, the illustrations demonstrate the structural knowledge leads to the misunderstandings for reasoning. 
In Figure~\ref{fig:case_study}.c, CoT fails to parse the relationship between two sales figures of entities, leading to cascading errors through the reasoning process. 
In Figure~\ref{fig:case_study}.d, CoT misinterprets ``0.75 bags per guest'' and ``1/4 of guests not attending'' as separate conditions, leading to double counting. 
SoT understands these relationships explicitly, preventing such misunderstandings.
In general, SoT facilitates the model to interpret relationships accurately by integrating with structured knowledge and language-specific knowledge, reducing errors caused by ambiguous expressions in different languages.

\section{Conclusion}
In this paper, we propose a Structured-of-Thought~(SoT) method for multilingual reasoning. By dynamically extracting entity-structured knowledge and language-specific structured knowledge, our method boost the ability to understand relationships in non-English questions for LLMs. Experimental results demonstrate that SoT achieves comparable performance on various LLMs, compared with several advanced methods. The analyses further indicates that SoT has both strong generalization capabilities and scalabilities, which can be integrated with other training-free strategies.

\section*{Limitations}
Existing multilingual benchmarks often rely on machine-translated text that introduces errors or includes expressions that are uncommon for native speakers. Due to the limitations of benchmarks, the cultural linguistic phenomena of native languages are uncertain. Thus, the impact of extracting language-specific knowledge may not be clearly reflected in existing benchmarks. The development of reasoning datasets for language-specific knowledge is urgent.
Moreover, in the first step, we utilize the Language Thinking Transformation to transfer the thinking pathway from the low-resource language to a high-resource language.
Generally speaking, English is the language chosen that performs best for various LLMs. However, some existing LLMs perform more prominently in other languages, which are trained with other languages as the core.
Therefore, selecting the target language for thinking transformation remains an urgent issue that needs to be addressed in the future.

\section*{Acknowledgments}
The research work descried in this paper has been supported by the Fundamental Research Funds for the Central Universities (2024JBZY019) and the National Nature Science Foundation of China (No. 62476023, 62406018, 62376019). The work is also supported by the Henan Provincial Science and Technology Research Project (No. 252102210102).
The authors would like to thank the anonymous reviewers for their valuable comments and suggestions to improve this paper.

\bibliography{acl_revised_latex}

\begin{thebibliography}{49}
\providecommand{\natexlab}[1]{#1}

\bibitem[{Ahuja et~al.(2023)Ahuja, Diddee, Hada, Ochieng, Ramesh, Jain, Nambi, Ganu, Segal, Ahmed, Bali, and Sitaram}]{ahuja-etal-2023-mega}
Kabir Ahuja, Harshita Diddee, Rishav Hada, Millicent Ochieng, Krithika Ramesh, Prachi Jain, Akshay Nambi, Tanuja Ganu, Sameer Segal, Mohamed Ahmed, Kalika Bali, and Sunayana Sitaram. 2023.
\newblock \href {https://doi.org/10.18653/v1/2023.emnlp-main.258} {{MEGA}: Multilingual evaluation of generative {AI}}.
\newblock In \emph{Proceedings of the 2023 Conference on Empirical Methods in Natural Language Processing}, pages 4232--4267, Singapore. Association for Computational Linguistics.

\bibitem[{Asai et~al.(2023)Asai, Kudugunta, Yu, Blevins, Gonen, Reid, Tsvetkov, Ruder, and Hajishirzi}]{asai2023buffetbenchmarkinglargelanguage}
Akari Asai, Sneha Kudugunta, Xinyan~Velocity Yu, Terra Blevins, Hila Gonen, Machel Reid, Yulia Tsvetkov, Sebastian Ruder, and Hannaneh Hajishirzi. 2023.
\newblock \href {https://arxiv.org/abs/2305.14857} {Buffet: Benchmarking large language models for few-shot cross-lingual transfer}.
\newblock \emph{Preprint}, arXiv:2305.14857.

\bibitem[{Bawden and Yvon(2023)}]{bawden-yvon-2023-investigating}
Rachel Bawden and Fran{\c{c}}ois Yvon. 2023.
\newblock \href {https://aclanthology.org/2023.eamt-1.16/} {Investigating the translation performance of a large multilingual language model: the case of {BLOOM}}.
\newblock In \emph{Proceedings of the 24th Annual Conference of the European Association for Machine Translation}, pages 157--170, Tampere, Finland. European Association for Machine Translation.

\bibitem[{Brown et~al.(2020)Brown, Mann, Ryder, Subbiah, Kaplan, Dhariwal, Neelakantan, Shyam, Sastry, Askell, Agarwal, Herbert-Voss, Krueger, Henighan, Child, Ramesh, Ziegler, Wu, Winter, Hesse, Chen, Sigler, Litwin, Gray, Chess, Clark, Berner, McCandlish, Radford, Sutskever, and Amodei}]{ICL}
Tom Brown, Benjamin Mann, Nick Ryder, Melanie Subbiah, Jared~D Kaplan, Prafulla Dhariwal, Arvind Neelakantan, Pranav Shyam, Girish Sastry, Amanda Askell, Sandhini Agarwal, Ariel Herbert-Voss, Gretchen Krueger, Tom Henighan, Rewon Child, Aditya Ramesh, Daniel Ziegler, Jeffrey Wu, Clemens Winter, Chris Hesse, Mark Chen, Eric Sigler, Mateusz Litwin, Scott Gray, Benjamin Chess, Jack Clark, Christopher Berner, Sam McCandlish, Alec Radford, Ilya Sutskever, and Dario Amodei. 2020.
\newblock \href {https://proceedings.neurips.cc/paper_files/paper/2020/file/1457c0d6bfcb4967418bfb8ac142f64a-Paper.pdf} {Language models are few-shot learners}.
\newblock In \emph{Advances in Neural Information Processing Systems}, volume~33, pages 1877--1901. Curran Associates, Inc.

\bibitem[{Chai et~al.(2025)Chai, Yang, Sun, Guo, Liu, Wang, Liang, Bai, Li, Peng et~al.}]{chai2025xcot}
Linzheng Chai, Jian Yang, Tao Sun, Hongcheng Guo, Jiaheng Liu, Bing Wang, Xinnian Liang, Jiaqi Bai, Tongliang Li, Qiyao Peng, et~al. 2025.
\newblock xcot: Cross-lingual instruction tuning for cross-lingual chain-of-thought reasoning.
\newblock In \emph{Proceedings of the AAAI Conference on Artificial Intelligence}, pages 23550--23558.

\bibitem[{Chen et~al.(2024)Chen, Zheng, Wu, Gong, Zhang, and Li}]{msvamp}
Nuo Chen, Zinan Zheng, Ning Wu, Ming Gong, Dongmei Zhang, and Jia Li. 2024.
\newblock \href {https://doi.org/10.18653/v1/2024.findings-emnlp.411} {Breaking language barriers in multilingual mathematical reasoning: Insights and observations}.
\newblock In \emph{Findings of the Association for Computational Linguistics: EMNLP 2024}, pages 7001--7016, Miami, Florida, USA. Association for Computational Linguistics.

\bibitem[{Chen et~al.(2023)Chen, Zheng, Wu, Shou, Gong, Song, Zhang, and Li}]{chen2023breaking}
Nuo Chen, Zinan Zheng, Ning Wu, Linjun Shou, Ming Gong, Yangqiu Song, Dongmei Zhang, and Jia Li. 2023.
\newblock \href {https://arxiv.org/abs/2310.20246} {Breaking language barriers in multilingual mathematical reasoning: Insights and observations}.
\newblock \emph{Preprint}, arXiv:2310.20246.

\bibitem[{Cheng et~al.(2024)Cheng, Ahmed, Willke, and Sun}]{cheng-etal-2024-structure}
Kewei Cheng, Nesreen~K. Ahmed, Theodore~L. Willke, and Yizhou Sun. 2024.
\newblock \href {https://doi.org/10.18653/v1/2024.emnlp-main.528} {Structure guided prompt: Instructing large language model in multi-step reasoning by exploring graph structure of the text}.
\newblock In \emph{Proceedings of the 2024 Conference on Empirical Methods in Natural Language Processing}, pages 9407--9430, Miami, Florida, USA. Association for Computational Linguistics.

\bibitem[{Chuang et~al.(2024)Chuang, Xie, Luo, Kim, Glass, and He}]{DOLA}
Yung{-}Sung Chuang, Yujia Xie, Hongyin Luo, Yoon Kim, James~R. Glass, and Pengcheng He. 2024.
\newblock \href {https://openreview.net/forum?id=Th6NyL07na} {Dola: Decoding by contrasting layers improves factuality in large language models}.
\newblock In \emph{The Twelfth International Conference on Learning Representations, {ICLR} 2024, Vienna, Austria, May 7-11, 2024}. OpenReview.net.

\bibitem[{Etxaniz et~al.(2024)Etxaniz, Azkune, Soroa, Lacalle, and Artetxe}]{etxaniz2024multilingual}
Julen Etxaniz, Gorka Azkune, Aitor Soroa, Oier Lacalle, and Mikel Artetxe. 2024.
\newblock Do multilingual language models think better in english?
\newblock In \emph{Proceedings of the 2024 Conference of the North American Chapter of the Association for Computational Linguistics: Human Language Technologies (Volume 2: Short Papers)}, pages 550--564.

\bibitem[{Fu et~al.(2023)Fu, Peng, Sabharwal, Clark, and Khot}]{fu2023complexitybased}
Yao Fu, Hao Peng, Ashish Sabharwal, Peter Clark, and Tushar Khot. 2023.
\newblock \href {https://openreview.net/forum?id=yf1icZHC-l9} {Complexity-based prompting for multi-step reasoning}.
\newblock In \emph{The Eleventh International Conference on Learning Representations}.

\bibitem[{Ghosh et~al.(2025)Ghosh, Datta, Saha, and Agarwal}]{ghosh2025multilingualmindsurvey}
Akash Ghosh, Debayan Datta, Sriparna Saha, and Chirag Agarwal. 2025.
\newblock \href {https://arxiv.org/abs/2502.09457} {The multilingual mind : A survey of multilingual reasoning in language models}.
\newblock \emph{Preprint}, arXiv:2502.09457.

\bibitem[{Hu et~al.(2025)Hu, Liu, Gao, Huang, Han, Feng, Deng, and Huang}]{hu-etal-2025-large-language}
Peng Hu, Sizhe Liu, Changjiang Gao, Xin Huang, Xue Han, Junlan Feng, Chao Deng, and Shujian Huang. 2025.
\newblock \href {https://aclanthology.org/2025.naacl-long.72/} {Large language models are cross-lingual knowledge-free reasoners}.
\newblock In \emph{Proceedings of the 2025 Conference of the Nations of the Americas Chapter of the Association for Computational Linguistics: Human Language Technologies (Volume 1: Long Papers)}, pages 1525--1542, Albuquerque, New Mexico. Association for Computational Linguistics.

\bibitem[{Huang et~al.(2025{\natexlab{a}})Huang, Mo, Chen, Guan, Yue, Wang, Xu, and Huang}]{huang2025boosting}
Chao Huang, Fengran Mo, Yufeng Chen, Changhao Guan, Zhenrui Yue, Xinyu Wang, Jinan Xu, and Kaiyu Huang. 2025{\natexlab{a}}.
\newblock Boosting data utilization for multilingual dense retrieval.
\newblock In \emph{Proceedings of the 2025 Conference on Empirical Methods in Natural Language Processing (EMNLP)}.

\bibitem[{Huang et~al.(2023)Huang, Tang, Zhang, Zhao, Song, Xia, and Wei}]{huang-etal-2023-languages}
Haoyang Huang, Tianyi Tang, Dongdong Zhang, Xin Zhao, Ting Song, Yan Xia, and Furu Wei. 2023.
\newblock \href {https://doi.org/10.18653/v1/2023.findings-emnlp.826} {Not all languages are created equal in {LLM}s: Improving multilingual capability by cross-lingual-thought prompting}.
\newblock In \emph{Findings of the Association for Computational Linguistics: EMNLP 2023}, pages 12365--12394, Singapore. Association for Computational Linguistics.

\bibitem[{Huang et~al.(2025{\natexlab{b}})Huang, Mo, Zhang, Li, Li, Zhang, Yi, Mao, Liu, Xu, Xu, Nie, and Liu}]{huang2025surveylargelanguagemodels}
Kaiyu Huang, Fengran Mo, Xinyu Zhang, Hongliang Li, You Li, Yuanchi Zhang, Weijian Yi, Yulong Mao, Jinchen Liu, Yuzhuang Xu, Jinan Xu, Jian-Yun Nie, and Yang Liu. 2025{\natexlab{b}}.
\newblock \href {https://arxiv.org/abs/2405.10936} {A survey on large language models with multilingualism: Recent advances and new frontiers}.
\newblock \emph{Preprint}, arXiv:2405.10936.

\bibitem[{Huang et~al.(2024)Huang, Zhu, Cheng, Li, and Yuan}]{huangmindmerger}
Zixian Huang, Wenhao Zhu, Gong Cheng, Lei Li, and Fei Yuan. 2024.
\newblock Mindmerger: Efficiently boosting llm reasoning in non-english languages.
\newblock In \emph{The Thirty-eighth Annual Conference on Neural Information Processing Systems}.

\bibitem[{Ji et~al.(2025)Ji, Li, Ye, Wu, Xu, Mo, and Zhang}]{DBLP:journals/corr/abs-2501-02497}
Yixin Ji, Juntao Li, Hai Ye, Kaixin Wu, Jia Xu, Linjian Mo, and Min Zhang. 2025.
\newblock \href {https://doi.org/10.48550/ARXIV.2501.02497} {Test-time computing: from system-1 thinking to system-2 thinking}.
\newblock \emph{CoRR}, abs/2501.02497.

\bibitem[{Jin et~al.(2024)Jin, Yu, Shu, Zhao, Hua, Meng, Zhang, and Du}]{jin-etal-2024-length}
Mingyu Jin, Qinkai Yu, Dong Shu, Haiyan Zhao, Wenyue Hua, Yanda Meng, Yongfeng Zhang, and Mengnan Du. 2024.
\newblock \href {https://doi.org/10.18653/v1/2024.findings-acl.108} {The impact of reasoning step length on large language models}.
\newblock In \emph{Findings of the Association for Computational Linguistics: ACL 2024}, pages 1830--1842, Bangkok, Thailand. Association for Computational Linguistics.

\bibitem[{Khot et~al.(2022)Khot, Trivedi, Finlayson, Fu, Richardson, Clark, and Sabharwal}]{khot2022decomposed}
Tushar Khot, Harsh Trivedi, Matthew Finlayson, Yao Fu, Kyle Richardson, Peter Clark, and Ashish Sabharwal. 2022.
\newblock Decomposed prompting: A modular approach for solving complex tasks.
\newblock \emph{arXiv preprint arXiv:2210.02406}.

\bibitem[{Kojima et~al.(2022)Kojima, Gu, Reid, Matsuo, and Iwasawa}]{kojima2022large}
Takeshi Kojima, Shixiang~Shane Gu, Machel Reid, Yutaka Matsuo, and Yusuke Iwasawa. 2022.
\newblock Large language models are zero-shot reasoners.
\newblock \emph{Advances in neural information processing systems}, 35:22199--22213.

\bibitem[{Koo and Kim(2025)}]{koo2025extracting}
Hamin Koo and Jaehyung Kim. 2025.
\newblock Extracting and emulsifying cultural explanation to improve multilingual capability of llms.
\newblock \emph{arXiv preprint arXiv:2503.05846}.

\bibitem[{Li et~al.(2024)Li, Wang, Zhang, and Zong}]{li-etal-2024-improving-context}
Chong Li, Shaonan Wang, Jiajun Zhang, and Chengqing Zong. 2024.
\newblock \href {https://doi.org/10.18653/v1/2024.naacl-long.445} {Improving in-context learning of multilingual generative language models with cross-lingual alignment}.
\newblock In \emph{Proceedings of the 2024 Conference of the North American Chapter of the Association for Computational Linguistics: Human Language Technologies (Volume 1: Long Papers)}, pages 8058--8076, Mexico City, Mexico. Association for Computational Linguistics.

\bibitem[{Li et~al.(2023)Li, Lin, Zhang, Fu, Chen, Lou, and Chen}]{li-etal-2023-making}
Yifei Li, Zeqi Lin, Shizhuo Zhang, Qiang Fu, Bei Chen, Jian-Guang Lou, and Weizhu Chen. 2023.
\newblock \href {https://doi.org/10.18653/v1/2023.acl-long.291} {Making language models better reasoners with step-aware verifier}.
\newblock In \emph{Proceedings of the 61st Annual Meeting of the Association for Computational Linguistics (Volume 1: Long Papers)}, pages 5315--5333, Toronto, Canada. Association for Computational Linguistics.

\bibitem[{Liu et~al.(2024)Liu, Zhang, Zhao, Luu, and Bing}]{liu2024translationneedstudysolving}
Chaoqun Liu, Wenxuan Zhang, Yiran Zhao, Anh~Tuan Luu, and Lidong Bing. 2024.
\newblock \href {https://arxiv.org/abs/2403.10258} {Is translation all you need? a study on solving multilingual tasks with large language models}.
\newblock \emph{Preprint}, arXiv:2403.10258.

\bibitem[{Lu et~al.(2024)Lu, Li, and Lam}]{lu2024dictionary}
Hongyuan Lu, Zixuan Li, and Wai Lam. 2024.
\newblock Dictionary insertion prompting for multilingual reasoning on multilingual large language models.
\newblock \emph{arXiv preprint arXiv:2411.01141}.

\bibitem[{Man et~al.(2024)Man, Huang, Zhang, Chen, Chen, and Xu}]{man-etal-2024-icl}
Zhibo Man, Kaiyu Huang, Yujie Zhang, Yuanmeng Chen, Yufeng Chen, and Jinan Xu. 2024.
\newblock \href {https://doi.org/10.18653/v1/2024.findings-emnlp.455} {{ICL}: Iterative continual learning for multi-domain neural machine translation}.
\newblock In \emph{Findings of the Association for Computational Linguistics: EMNLP 2024}, pages 7732--7743, Miami, Florida, USA. Association for Computational Linguistics.

\bibitem[{Man et~al.(2023)Man, Huang, Zhang, Li, Chen, Chen, and Xu}]{man2023wdsrl}
Zhibo Man, Zengcheng Huang, Yujie Zhang, Yu~Li, Yuanmeng Chen, Yufeng Chen, and Jinan Xu. 2023.
\newblock Wdsrl: Multi-domain neural machine translation with word-level domain-sensitive representation learning.
\newblock \emph{IEEE/ACM Transactions on Audio, Speech, and Language Processing}, 32:577--590.

\bibitem[{Mo et~al.(2025{\natexlab{a}})Mo, Gao, Meng, Liu, Wu, Mao, Wang, Chen, Li, Li et~al.}]{mo2025uniconv}
Fengran Mo, Yifan Gao, Chuan Meng, Xin Liu, Zhuofeng Wu, Kelong Mao, Zhengyang Wang, Pei Chen, Zheng Li, Xian Li, et~al. 2025{\natexlab{a}}.
\newblock Uniconv: Unifying retrieval and response generation for large language models in conversations.
\newblock In \emph{Proceedings of the 63rd Annual Meeting of the Association for Computational Linguistics (Volume 1: Long Papers)}, pages 6936--6949.

\bibitem[{Mo et~al.(2025{\natexlab{b}})Mo, Mao, Zhao, Qian, Chen, Cheng, Li, Zhu, Dou, and Nie}]{mo2024survey}
Fengran Mo, Kelong Mao, Ziliang Zhao, Hongjin Qian, Haonan Chen, Yiruo Cheng, Xiaoxi Li, Yutao Zhu, Zhicheng Dou, and Jian-Yun Nie. 2025{\natexlab{b}}.
\newblock A survey of conversational search.
\newblock \emph{ACM Transactions on Information Systems (TOIS)}.

\bibitem[{Ponti et~al.(2020)Ponti, Glava{\v{s}}, Majewska, Liu, Vuli{\'c}, and Korhonen}]{xcopa}
Edoardo~Maria Ponti, Goran Glava{\v{s}}, Olga Majewska, Qianchu Liu, Ivan Vuli{\'c}, and Anna Korhonen. 2020.
\newblock \href {https://doi.org/10.18653/v1/2020.emnlp-main.185} {{XCOPA}: A multilingual dataset for causal commonsense reasoning}.
\newblock In \emph{Proceedings of the 2020 Conference on Empirical Methods in Natural Language Processing (EMNLP)}, pages 2362--2376, Online. Association for Computational Linguistics.

\bibitem[{Qin et~al.(2023)Qin, Chen, Wei, Huang, and Che}]{qin-etal-2023-cross}
Libo Qin, Qiguang Chen, Fuxuan Wei, Shijue Huang, and Wanxiang Che. 2023.
\newblock \href {https://doi.org/10.18653/v1/2023.emnlp-main.163} {Cross-lingual prompting: Improving zero-shot chain-of-thought reasoning across languages}.
\newblock In \emph{Proceedings of the 2023 Conference on Empirical Methods in Natural Language Processing}, pages 2695--2709, Singapore. Association for Computational Linguistics.

\bibitem[{Radford et~al.(2019)Radford, Wu, Child, Luan, Amodei, Sutskever et~al.}]{radford2019language}
Alec Radford, Jeffrey Wu, Rewon Child, David Luan, Dario Amodei, Ilya Sutskever, et~al. 2019.
\newblock Language models are unsupervised multitask learners.
\newblock \emph{OpenAI blog}, 1(8):9.

\bibitem[{She et~al.(2024)She, Zou, Huang, Zhu, Liu, Geng, and Chen}]{she-etal-2024-mapo}
Shuaijie She, Wei Zou, Shujian Huang, Wenhao Zhu, Xiang Liu, Xiang Geng, and Jiajun Chen. 2024.
\newblock \href {https://doi.org/10.18653/v1/2024.acl-long.539} {{MAPO}: Advancing multilingual reasoning through multilingual-alignment-as-preference optimization}.
\newblock In \emph{Proceedings of the 62nd Annual Meeting of the Association for Computational Linguistics (Volume 1: Long Papers)}, pages 10015--10027, Bangkok, Thailand. Association for Computational Linguistics.

\bibitem[{Shi et~al.(2023)Shi, Suzgun, Freitag, Wang, Srivats, Vosoughi, Chung, Tay, Ruder, Zhou, Das, and Wei}]{shi2023language}
Freda Shi, Mirac Suzgun, Markus Freitag, Xuezhi Wang, Suraj Srivats, Soroush Vosoughi, Hyung~Won Chung, Yi~Tay, Sebastian Ruder, Denny Zhou, Dipanjan Das, and Jason Wei. 2023.
\newblock \href {https://openreview.net/forum?id=fR3wGCk-IXp} {Language models are multilingual chain-of-thought reasoners}.
\newblock In \emph{The Eleventh International Conference on Learning Representations}.

\bibitem[{Wang et~al.(2024)Wang, Yang, Hu, Shen, Zhang, and Gu}]{wang2024know}
Junjie Wang, Dan Yang, Binbin Hu, Yue Shen, Wen Zhang, and Jinjie Gu. 2024.
\newblock Know your needs better: Towards structured understanding of marketer demands with analogical reasoning augmented llms.
\newblock In \emph{Proceedings of the 30th ACM SIGKDD Conference on Knowledge Discovery and Data Mining}, pages 5860--5871.

\bibitem[{Wei et~al.(2022)Wei, Wang, Schuurmans, Bosma, ichter, Xia, Chi, Le, and Zhou}]{NEURIPS2022_9d560961}
Jason Wei, Xuezhi Wang, Dale Schuurmans, Maarten Bosma, brian ichter, Fei Xia, Ed~Chi, Quoc~V Le, and Denny Zhou. 2022.
\newblock \href {https://proceedings.neurips.cc/paper_files/paper/2022/file/9d5609613524ecf4f15af0f7b31abca4-Paper-Conference.pdf} {Chain-of-thought prompting elicits reasoning in large language models}.
\newblock In \emph{Advances in Neural Information Processing Systems}, volume~35, pages 24824--24837. Curran Associates, Inc.

\bibitem[{Zhang et~al.(2024{\natexlab{a}})Zhang, Mo, Wang, and Liu}]{zhang2024blind}
Jinghan Zhang, Fengran Mo, Xiting Wang, and Kunpeng Liu. 2024{\natexlab{a}}.
\newblock Blind spot navigation in llm reasoning with thought space explorer.
\newblock \emph{arXiv preprint arXiv:2410.24155}.

\bibitem[{Zhang et~al.(2025{\natexlab{a}})Zhang, Wang, Mo, Zhou, Gao, and Liu}]{zhang2025entropy}
Jinghan Zhang, Xiting Wang, Fengran Mo, Yeyang Zhou, Wanfu Gao, and Kunpeng Liu. 2025{\natexlab{a}}.
\newblock Entropy-based exploration conduction for multi-step reasoning.
\newblock \emph{arXiv preprint arXiv:2503.15848}.

\bibitem[{Zhang et~al.(2025{\natexlab{b}})Zhang, Wang, Ren, Jiang, Wang, and Liu}]{zhang2025ratt}
Jinghan Zhang, Xiting Wang, Weijieying Ren, Lu~Jiang, Dongjie Wang, and Kunpeng Liu. 2025{\natexlab{b}}.
\newblock Ratt: A thought structure for coherent and correct llm reasoning.
\newblock In \emph{Proceedings of the AAAI Conference on Artificial Intelligence}, volume~39, pages 26733--26741.

\bibitem[{Zhang et~al.(2024{\natexlab{b}})Zhang, Gautam, Wang, Alabi, Shen, Klakow, and Mosbach}]{zhang-etal-2024-impact}
Miaoran Zhang, Vagrant Gautam, Mingyang Wang, Jesujoba Alabi, Xiaoyu Shen, Dietrich Klakow, and Marius Mosbach. 2024{\natexlab{b}}.
\newblock \href {https://doi.org/10.18653/v1/2024.findings-acl.438} {The impact of demonstrations on multilingual in-context learning: A multidimensional analysis}.
\newblock In \emph{Findings of the Association for Computational Linguistics: ACL 2024}, pages 7342--7371, Bangkok, Thailand. Association for Computational Linguistics.

\bibitem[{Zhang et~al.(2023)Zhang, Cahyawijaya, Cruz, Winata, and Aji}]{zhang-etal-2023-multilingual}
Ruochen Zhang, Samuel Cahyawijaya, Jan Christian~Blaise Cruz, Genta Winata, and Alham~Fikri Aji. 2023.
\newblock \href {https://doi.org/10.18653/v1/2023.emnlp-main.774} {Multilingual large language models are not (yet) code-switchers}.
\newblock In \emph{Proceedings of the 2023 Conference on Empirical Methods in Natural Language Processing}, pages 12567--12582, Singapore. Association for Computational Linguistics.

\bibitem[{Zhang et~al.(2025{\natexlab{c}})Zhang, Liang, Meng, Zhang, Chen, Xu, and Zhou}]{zhang2025lessbetterefficientmultilingual}
Xue Zhang, Yunlong Liang, Fandong Meng, Songming Zhang, Yufeng Chen, Jinan Xu, and Jie Zhou. 2025{\natexlab{c}}.
\newblock \href {https://arxiv.org/abs/2505.22582} {Less, but better: Efficient multilingual expansion for llms via layer-wise mixture-of-experts}.
\newblock \emph{Preprint}, arXiv:2505.22582.

\bibitem[{Zheng et~al.(2023)Zheng, Liu, Xie, Li, and Li}]{zheng2023progressive}
Chuanyang Zheng, Zhengying Liu, Enze Xie, Zhenguo Li, and Yu~Li. 2023.
\newblock Progressive-hint prompting improves reasoning in large language models.
\newblock \emph{arXiv preprint arXiv:2304.09797}.

\bibitem[{Zhou et~al.(2022)Zhou, Sch{\"a}rli, Hou, Wei, Scales, Wang, Schuurmans, Cui, Bousquet, Le et~al.}]{zhou2022least}
Denny Zhou, Nathanael Sch{\"a}rli, Le~Hou, Jason Wei, Nathan Scales, Xuezhi Wang, Dale Schuurmans, Claire Cui, Olivier Bousquet, Quoc Le, et~al. 2022.
\newblock Least-to-most prompting enables complex reasoning in large language models.
\newblock \emph{arXiv preprint arXiv:2205.10625}.

\bibitem[{Zhu et~al.(2024{\natexlab{a}})Zhu, Huang, Yuan, She, Chen, and Birch}]{zhu-etal-2024-question}
Wenhao Zhu, Shujian Huang, Fei Yuan, Shuaijie She, Jiajun Chen, and Alexandra Birch. 2024{\natexlab{a}}.
\newblock \href {https://doi.org/10.18653/v1/2024.findings-acl.498} {Question translation training for better multilingual reasoning}.
\newblock In \emph{Findings of the Association for Computational Linguistics: ACL 2024}, pages 8411--8423, Bangkok, Thailand. Association for Computational Linguistics.

\bibitem[{Zhu et~al.(2024{\natexlab{b}})Zhu, Liu, Dong, Xu, Huang, Kong, Chen, and Li}]{zhu2024multilingualmachinetranslationlarge}
Wenhao Zhu, Hongyi Liu, Qingxiu Dong, Jingjing Xu, Shujian Huang, Lingpeng Kong, Jiajun Chen, and Lei Li. 2024{\natexlab{b}}.
\newblock \href {https://arxiv.org/abs/2304.04675} {Multilingual machine translation with large language models: Empirical results and analysis}.
\newblock \emph{Preprint}, arXiv:2304.04675.

\bibitem[{Zhu et~al.(2024{\natexlab{c}})Zhu, Liu, Huang, She, Wendler, and Chen}]{SL-D}
Wenhao Zhu, Sizhe Liu, Shujian Huang, Shuaijie She, Chris Wendler, and Jiajun Chen. 2024{\natexlab{c}}.
\newblock \href {https://aclanthology.org/2024.findings-emnlp.512} {Multilingual contrastive decoding via language-agnostic layers skipping}.
\newblock In \emph{Findings of the Association for Computational Linguistics: {EMNLP} 2024, Miami, Florida, USA, November 12-16, 2024}, pages 8775--8782. Association for Computational Linguistics.

\bibitem[{Zhu et~al.(2023)Zhu, Lv, Dong, Yuan, Xu, Huang, Kong, Chen, and Li}]{zhu2023extrapolatinglargelanguagemodels}
Wenhao Zhu, Yunzhe Lv, Qingxiu Dong, Fei Yuan, Jingjing Xu, Shujian Huang, Lingpeng Kong, Jiajun Chen, and Lei Li. 2023.
\newblock \href {https://arxiv.org/abs/2308.04948} {Extrapolating large language models to non-english by aligning languages}.
\newblock \emph{Preprint}, arXiv:2308.04948.

\end{thebibliography}

\clearpage
\appendix

\section{Expermental Details}
\label{sec:appendix}

\subsection{Dataset Details}
\label{sec:data_details}
\paragraph{MGSM (Multilingual Grade School Math).} 
MGSM \citep{shi2023language} is a benchmark of multilingual elementary school math reasoning problems. The dataset is translated from the GSM8K dataset and contains 11 different languages, which aims to evaluate the ability of models to solve math problems in a multilingual environment.

\paragraph{MSVAMP (Multilingual Semantic Value Math Problems).} MSVAMP \citep{msvamp} is a math problem dataset focusing on multilingual semantic reasoning, designed to evaluate the mathematical reasoning and semantic understanding ability of models in different languages. The dataset contains math problems in multiple languages, emphasizing the understanding of quantity, units, and measurement words.

\paragraph{XCOPA (Cross-lingual Choice of Plausible Alternatives).} XCOPA \cite{xcopa} is a benchmark for multilingual commonsense reasoning tasks. The questions involve reasoning scenarios in multiple cultural backgrounds and support more than ten languages, including English, Arabic, Chinese, Spanish, French, German, Russian, etc. The benchmark aims to test cross-language reasoning capabilities and the adaptability of models to different cultural backgrounds.

\subsection{Baselines}
\label{app_baselines}
We compare our method with various representative baselines in multilingual reasoning. 
A branch of baselines is the training-free methods, listed as follows:
\begin{itemize}
    \item Direct:~Only the most basic prompt strategy (such as "Let’s solve the following problem") is used without any additional prompt strategy.
    \item Few-Shot:~We use three examples along with instructions as input to demonstrate the problem-solving steps to the LLMs.
    \item CoT~\cite{NEURIPS2022_9d560961}:~The model is instructed to reason in English using the phrase "Let’s think step by step in English.
    \item DoLa~\cite{DOLA}:~DOLA contrasts logits between early and later layers to emphasize factual knowledge from higher layers, reducing hallucinations and improving the truthfulness of the generated output.
    \item SL-D~\cite{SL-D}:~By skipping language-agnostic lower layers and contrasting early exit outputs with final outputs, the model leverages more accurate amateur logits to enhance multilingual reasoning.
    \item DIP~\cite{lu2024dictionary}:~DIP inserts dictionary-based English counterparts into non-English queries, enabling LLMs to better translate and reason in English, all while remaining computationally lightweight.
    \item CLP~\cite{qin-etal-2023-cross}:~CLP aligns representations between source and target languages and uses task-specific solver prompting to guide reasoning.
    \item EMCEI~\cite{koo2025extracting}:~EMCEI extractes relevant cultural context from LLMs and combines it with reasoning steps to select the most contextually appropriate response using an LLM-as-Judge mechanism.
\end{itemize}
Another branch of baselines is the post-training methods, listed as follows:
\begin{itemize}
    \item xCoT~\cite{chai2025xcot}:~XCOT fine-tunes LLMs on multilingual instruction data by using cross-lingual CoT reasoning.
    \item QAlign~\cite{zhu-etal-2024-question}:~QAlign fine-tunes models to translate reasoning questions into English for better alignment with English instruction data, followed by response alignment to enhance performance in non-English languages.
    \item MindMerger~\cite{huangmindmerger}:~MindMerger embeds external language understanding capabilities into LLMs and training them to collaboratively use both internal and external knowledge through a two-step training process.
    \item MAPO~\cite{she-etal-2024-mapo}:~MAPO aligns non-dominant language reasoning with a dominant language through preference estimation and optimization techniques like Proximal Policy Optimization (PPO) and Direct Preference Optimization (DPO).
\end{itemize}

\section{Case Study}
\label{sec:case_study}
In this section, we showcase additional examples of responses generated by \textbf{\texttt{gpt-3.5-turbo}}, utilizing both CoT and SoT method for each task. 

\subsection{Examples in MSVAMP}
\label{sec:ex_msvamp}

\begin{tcolorbox}[colback=lightgray!10,  
                  colframe=black,  
                  arc=2mm,         
                  boxrule=0.5pt,   
                  width=\linewidth, 
                  boxsep=2pt, 
                  left=4pt, right=4pt,  
                  top=4pt, bottom=4pt
                  ] 
\small
\textbf{\underline{Question:}} ``瑞秋有52棵苹果树。她从其中一棵树上摘了2个苹果。现在这棵树上还剩下7个苹果。这棵树最初有多少个苹果？''

\noindent\textbf{\underline{CoT Answer:}}

Step 1: Calculate the total number of apples picked from all trees: 52 trees * 2 apples = 104 apples

Step 2: Calculate the total number of apples left on all trees: 52 trees * 7 apples = 364 apples

Step 3: Calculate the total number of apples on the tree before picking:
Total - Picked - Left = Initial

Initial = 364 - 104 = 260

Answer: 初始苹果数为260个。

\noindent\textbf{\underline{SoT Answer:}}

\textbf{1. Problem-solving process using English thinking:}

Determine how many apples the tree originally had.

Subtract the apples Rachel picked from the total apples on the tree.

\textbf{2. Extract relationships between numbers in order:}

Total number of apple trees: 52

Apples picked from one tree: 2

Apples left on the same tree: 7

\textbf{3. Language-specific knowledge:}

No specific language-specific knowledge is needed here.

\textbf{4. Calculation:}
If the tree has 7 apples remaining after Rachel picked 2 apples, the tree must have originally had:

7 (remaining apples) + 2 (apples picked) = 9( original number of apples).

\textbf{Final Answer:} 这棵树最初有 9 个苹果。
\end{tcolorbox}














\begin{tcolorbox}[colback=lightgray!10,  
                  colframe=black,  
                  arc=2mm,         
                  boxrule=0.5pt,   
                  width=\linewidth, 
                  boxsep=2pt, 
                  left=4pt, right=4pt,  
                  top=4pt, bottom=4pt
                  ] 
\small

\noindent\textbf{\underline{Question:}} ``面包师傅做了61个糕点和167个蛋糕。如果他卖掉了108个蛋糕和44个糕点，面包师傅还剩下多少个蛋糕？''

\noindent\textbf{\underline{CoT Answer:}}

Step 1: Calculate the total number of pastries and cakes baker made originally:\\61 (pastries) + 167 (cakes) = 228

Step 2: Calculate the total number of pastries and cakes baker sold:\\108 (cakes) + 44 (pastries) = 152

Step 3: Calculate the total number of pastries and cakes baker would still have:\\228 (total) - 152 (sold) = 76

Answer: 面包师傅还会有76个蛋糕。

\noindent\textbf{\underline{SoT Answer:}}

\textbf{1. Problem-solving process using English thinking:}

Determine how many cakes the baker has left after selling some.

Subtract the cakes he sold from the total number of cakes he originally made.

\textbf{2. Extract relationships between numbers:}

Number of pastries made: 61

Number of cakes made: 167

Number of cakes sold: 108

Number of pastries sold: 44

\textbf{3. Language-specific knowledge:}

``糕点'' refers to pastries, and ``蛋糕'' refers to cakes.

\textbf{4. Calculation:}

The baker made 167 cakes and sold 108 of them.

The number of cakes the baker still has is:

167(original cakes) − 108 (cakes sold) = 59  (cakes remaining).

\textbf{Final Answer: }The baker still has 59 cakes.
\end{tcolorbox}

\subsection{Examples in XCOPA}
\label{sec:ex_xcopa}

\begin{tcolorbox}[colback=lightgray!10,  
                  colframe=black,  
                  arc=2mm,         
                  boxrule=0.5pt,   
                  width=\linewidth, 
                  boxsep=2pt, 
                  left=4pt, right=4pt,  
                  top=4pt, bottom=4pt
                  ] 
\small
\textbf{\underline{Question:}} 

\textbf{premise:} ``我讲话停了下来了。''

\textbf{choice 0: }``我失声了。''

\textbf{choice 1:} ``我气喘吁吁。''

\textbf{question:} ``cause'',

\noindent\textbf{\underline{CoT Answer:}}

To determine which choice is more relevant and aligned with the premise and question, let's analyze both options:

Choice 0: 我失声了。 (I lost my voice.)\\- This choice suggests that the speaker stopped talking because they lost their voice, which is a plausible cause for the situation described in the premise.

Choice 1: 我气喘吁吁。 (I am panting.)\\- This choice suggests that the speaker stopped talking because they are out of breath, which may not be directly related to the premise of the speaker's speech stopping.

Based on the analysis, Choice 0: 我失声了。 (I lost my voice.) is more relevant and aligned with the premise and question. Therefore, the final answer is ``0'' for Choice 0.

\noindent\textbf{\underline{SoT Answer:}}

\textbf{1. Premise, question, and choices in English:}

Premise: ``I stopped speaking.''

Question: What caused it?

Choice 0: ``I lost my voice.''

Choice 1: ``I was out of breath.''

\textbf{2. Extract relationships from the premise:}

Subject: ``I''

Action: ``stopped speaking''

This suggests a sudden or involuntary interruption in speech.

\textbf{3. Language-specific analysis of choices:}

In Chinese, ``我失声了'' means the speaker can no longer speak or has become mute or hoarse.

\textbf{4. Comparison:}

Choice 0 : This suggests a more sudden, possibly longer-term vocal issue, like losing the ability to speak completely.

Choice 1 : It fits better because it indicates a temporary, immediate physical condition that would logically cause someone to pause speaking, especially if they were: out of breath from exercise or running.

\textbf{Final Answer: }The more plausible cause of stopping speaking is out of breath. So the final answer is: 1.
\end{tcolorbox}

\section{Supplementary Results}
\label{sec:all_results}



In this study, we investigate the effectiveness of SoT on three multilingual reasoning benchmarks using four different LLMs.
Moreover, we also integrate other training-free methods into SoT.
All results demonstrate that SoT achieves good performance for multilingual reasoning with a strong generalization ability, as shown in Table~\ref{tab:supresults_mgsm}, Table~\ref{tab:supresults_msvamp} and Table~\ref{tab:supresults_xcopa}.

\begin{table*}[!t]
    \centering
    \small
    \begin{tabular}{lcccccccccccc}
        \toprule
        \multirow{2}{*}{\textbf{Methods}}   & \multicolumn{11}{c}{\textbf{Language}} \\
        \cmidrule(lr){2-12}
         \textbf{ } & \textbf{En} & \textbf{Sw} & \textbf{Ja} & \textbf{Be} & \textbf{Th} & \textbf{Te} & \textbf{Ru} & \textbf{Zh} & \textbf{De} & \textbf{Es} & \textbf{Fr} & \textbf{Avg.} \\
        \midrule
        (\textbf{\texttt{training-free}}) & \multicolumn{12}{c}{(\texttt{gpt-3.5-turbo})}\\
        Direct & & 37.3 \\
        DoLa & 72.2 & 57.2 & 58.0 & 43.4 & 52.2 & 17.6 & 62.4 & 62.2 & 60.0 & 68.2 & 58.8 & 55.7 \\
        DIP & 70.2 & 55.2 & 58.8 & 54.6 & 51.6 & 19.4 & 65.7 & 62.8 & 61.4 & 69.8 & 59.2 & 57.2 \\
        CLP & 73.2 & 55.8 & 59.4 & 56.0 & 53.6 & 28.0 & 66.2 & 64.6 & 64.8 & 71.4 & 60.0 & 59.4 \\
        EMCEI& 73.0 & 59.2 & 60.2 & 55.8 & 54.2 & 26.8 & 63.4 & 63.0 & 62.8 & 70.4 & 59.8 & 59.0 \\
        \hdashline
        SoT~(\textit{Ours}) & 74.4 & 62.0 & \textbf{65.2} & 61.2 & 56.0 & 34.0 & 67.6 & 67.2 & 66.8 & 72.8 & 63.2 & 62.8\\
        ~~+3-shot & 74.0 & \textbf{66.4} & 63.6 & \textbf{63.6} & \textbf{60.4} & 36.0 & \textbf{70.4} & \textbf{69.2} & \textbf{70.4} & 74.0 & \textbf{65.6} & \textbf{64.9} \\
        ~~+CoT & \textbf{75.2} & 64.8 & \textbf{65.2} & 57.6 & 55.6 & \textbf{38.4} & 69.2 & 65.6 & 68.0 & \textbf{76.4} & 64.0 & 63.6 \\
        \midrule
        \midrule
        (\textbf{\texttt{training-free}}) & \multicolumn{12}{c}{(\texttt{Qwen2.5-32B-Instruct})}\\
        Direct  & 87.2 & 53.6 & 84.2 & 85.6 & 82.4 & 82.4 & 86.8 & 81.2 & 81.2 & 75.6 & 57.2 & 77.9  \\
        DoLa & 85.0 & 45.2 & 71.6 & 80.8 & 69.1 & 62.0 & 77.2 & 82.6 & 76.2 & 73.6 & 53.2 & 70.6 \\
        SL-D & 85.8 & 57.2 & 82.4 & 83.4 & 78.3 & 80.2 & 86.4 & 83.0 & 82.8 & 76.6 & 62.0 & 78.0 \\
        DIP & 85.8 & 52.6 & 82.0 & 81.2 & 75.2 & 74.2 & 83.0 & 83.0 & 80.4 & 78.2 & 54.2 & 75.4 \\
        CLP & 86.0 & 53.6 & 81.4 & 84.2 & 78.2 & 77.2 & 84.2 & 83.8 & 81.4 & 77.2 & 58.2 & 76.9 \\
        EMCEI & 85.4 & 52.8 & 81.0 & 83.8 & 83.0 & 78.0 & 84.0 & 82.4 & 81.6 & 78.0 & 62.6 & 77.5 \\
        \hdashline
        SoT~(\textit{Ours}) & 87.2 & \textbf{67.2} & 86.0 & 86.0 & 85.4 & 87.4 & 88.8 & \textbf{84.4} & 86.4 & \textbf{78.4} & 64.8 & 82.0 \\
        ~~+3-shot & 87.2 & 66.8 & 86.2 & \textbf{87.6} & 86.4 & 88.0 & 88.4 & 82.4 & 86.4 & 78.0 & 66.8 & 82.2 \\
        ~~+CoT & \textbf{87.8} & 63.6 & \textbf{87.2} & 86.8 & \textbf{86.8} & \textbf{88.8} & \textbf{89.2} & 84.0 & \textbf{86.8} & 77.6 &\textbf{ 68.8} & \textbf{82.5} \\
        \midrule
        (\textbf{\texttt{post-training}}) \\
        xCoT & 86.6 & 58.4 & 83.2 & 82.4 & 81.4 & 27.4 & 80.8 & 87.2 & 81.0 & 81.2 & 82.0 & 75.6 \\
        QAlign & 86.4 & 58.0 & 80.0 & 81.2 & 84.0 & 29.2 & 85.6 & 86.2 & 81.6 & 82.0 & 81.6 & 76.0 \\
        MindMerger & 87.0 & 69.2 & 83.0 & 84.8 & 85.6 & 38.0 & 88.0 & 88.0 & 82.5 & 82.4 & 82.8 & 79.2 \\
        MAPO & 87.0 & 61.6 & 83.2 & 86.0 & 83.0 & 35.2 & 86.2 & 89.8 & 83.0 & 83.8 & 83.2 & 78.4 \\
        
        \bottomrule
    \end{tabular}%
     \caption{Supplementary results~(\%) of mathematical reasoning on MGSM using \textbf{\texttt{gpt-3.5-turbo}} and \textbf{\texttt{Qwen2.5-32B-Instruct}}.}
    \label{tab:supresults_mgsm}
\end{table*}

\begin{table*}[!t]
    \centering
    \small
    \begin{tabular}{lccccccccccc}
        \toprule
        \multirow{2}{*}{\textbf{Methods}}   & \multicolumn{11}{c}{\textbf{Language}} \\
        \cmidrule(lr){2-11}
         \textbf{ } & \textbf{En} & \textbf{Sw} & \textbf{Ja} & \textbf{Be} & \textbf{Th} & \textbf{Ru} & \textbf{Zh} & \textbf{De} & \textbf{Es} & \textbf{Fr} & \textbf{Avg.} \\
        \midrule
        (\textbf{\texttt{training-free}}) & \multicolumn{11}{c}{(\texttt{gpt-3.5-turbo})}\\
        Direct & 77.0 & 68.1 & 68.4 & 48.7 & 61.8 & 74.3 & 68.0 & 73.4 & 73.3 & 73.4 & 68.6 \\
        DoLa &  76.4 & 61.2 & 62.4 & 49.0 & 61.2 & 68.7 & 69.4 & 68.6 & 70.1 & 69.5 & 65.7 \\
        DIP & 70.0 & 68.4 & 69.8 & 50.5 & 64.0 & 69.4 & 75.2 & 75.8 & 73.0 & 74.5 & 69.1 \\
        CLP & 78.8 & 68.7 & 70.4 & 52.2 & 68.2 & 72.0 & 76.6 & 74.6 & 76.5 & 77.1 & 71.5 \\
        EMCEI & 73.8 & 69.0 & 70.8 & 52.0 & 66.5 & 70.6 & 74.2 & 73.6 & 74.3 & 76.3 & 70.1 \\
        \hdashline
         SoT~(\textit{Ours}) & 81.8 & 75.4 & 80.2 & 63.6 & 72.8 & \textbf{79.2} & 80.4 & 80.0 & \textbf{83.0} & 80.4 & 77.7 \\
         ~~+3-shot & 82.0 & 76.4 & 79.6 & \textbf{66.0} & 74.2 & \textbf{79.2} & 80.6 & 80.6 & 81.8 & 78.2 & 77.9 \\
         ~~+CoT & \textbf{82.4} & \textbf{76.6} & \textbf{81.0} & 64.2 & \textbf{74.4} & 78.4 & \textbf{81.4} & \textbf{81.6} & 81.4 & \textbf{81.0} & \textbf{78.2} \\
        \midrule
        \midrule
        (\textbf{\texttt{training-free}}) & \multicolumn{11}{c}{(\texttt{Qwen2.5-32B-Instruct})}\\
        Direct  & 89.8 & 39.4 & 69.2 & 55.0 & 65.4 & 74.6 & 81.8 & 77.8 & 83.2 & 82.2 & 71.8 \\
        DoLa & 85.2 & 48.3 & 76.1 & 68.2 & 71.9 & 87.3 & 83.9 & 75.7 & 81.3 & 78.4 & 75.6 \\
        SL-D & 88.2 & 54.5 & 79.4 & 81.4 & 83.8 & 88.5 & 86.7 & 81.2 & 85.2 & 82.2 & 81.1 \\
        DIP & 88.2 & 52.2 & 82.2 & 77.2 & 72.3 & 88.1 & 87.3 & 82.7 & 83.6 & 85.8 & 80.0 \\
        CLP & 90.8 & 53.0 & 82.6 & 73.1 & 78.2 & 88.2 & 86.8 & 83.2 & 83.3 & 89.1 & 80.8 \\
        EMCEI & 91.2 & 58.3 & 83.0 & 76.3 & 76.3 & 85.5 & 88.3 & 85.6 & 84.9 & 86.4 & 81.6 \\
        \hdashline
        SoT~(\textit{Ours}) & \textbf{93.8} & \textbf{87.4} & 89.8 & \textbf{84.8} & 87.0 & \textbf{90.8} & 91.8 & 91.8 & 92.6 & 93.2 & 90.3  \\
        ~~+3-shot & 93.7 & 86.4 & 91.0 & 83.1 & \textbf{87.6} & 90.2 & \textbf{93.6} &\textbf{93.4} & 93.8 & \textbf{93.8} & \textbf{90.7}  \\
        ~~+CoT & 94.2 & 87.0 & \textbf{92.0} & 83.6 & 86.4 & 89.8 & 91.4 & \textbf{93.4} & \textbf{94.4} & 93.6 & 90.6 \\
        \midrule
        (\textbf{\texttt{post-training}}) \\
        xCoT & 90.3 & 75.2 & 81.5 & 74.9 & 75.4 & 85.0 & 85.5 & 82.8 & 85.3 & 89.0 & 82.5 \\
        QAlign & 90.7 & 72.8 & 85.5 & 75.3 & 77.5 & 88.0 & 83.8 & 87.2 & 89.6 & 89.4 & 84.0 \\
        MindMerger & 91.5 & 77.0 & 85.8 & 78.5 & 78.2 & 87.1 & 86.8 & 88.5 & 90.2 & 91.3 & 85.5 \\
        MAPO & 91.9 & 71.1 & 86.0 & 74.0 & 79.1 & 82.5 & 86.3 & 85.6 & 88.4 & 89.4 & 83.4 \\
        
    \bottomrule
    \end{tabular}%
     \caption{Supplementary results~(\%) of mathematical reasoning on MSVAMP using \textbf{\texttt{gpt-3.5-turbo}} and \textbf{\texttt{Qwen2.5-32B-Instruct}}.}
    \label{tab:supresults_msvamp}
\end{table*}

\begin{table*}[!t]
    \centering
    \small
    \begin{tabular}{lcccccccccccc}
        \toprule
        \multirow{2}{*}{\textbf{Methods}}   & \multicolumn{11}{c}{\textbf{Language}} \\
        \cmidrule(lr){2-12}
         \textbf{ } & \textbf{Et} & \textbf{Ht} & \textbf{Id} & \textbf{It} & \textbf{Qu} & \textbf{Sw} & \textbf{Ta} & \textbf{Th} & \textbf{Tr} & \textbf{Vi} & \textbf{Zh} & \textbf{Avg.} \\
        \midrule
(\textbf{\texttt{training-free}}) & \multicolumn{12}{c}{(\texttt{DeepSeek-R1-7B})}\\
        Direct & 19.6 & 20.6 & 15.8 & 11.0 & 19.6 & 16.0 & 16.0 & 11.0 & 12.2 & 12.8 & 10.6 & 15.0\\
        DoLa & 23.4 & 34.0 & 48.4 & 29.6 & 30.2 & 35.2 & 32.0 & 43.8 & 44.6 & 49.4 & 67.0 & 39.8\\
        SL-D & 35.8 & 40.4 & 51.8 & 43.2 & 34.4 & 39.6 & 45.2 & 51.0 & 47.8 & 50.8 & 69.2 & 46.3 \\
        DIP & 33.8 & 40.2 & 51.4 & 50.4 & 41.2 & 47.2 & 47.2 & 43.9 & 45.8 & 50.0 & 66.0 & 47.0\\
        CLP & 30.0 & 43.6 & 54.8 & 48.4 & 42.0 & 43.0 & 49.4 & 51.2 & 52.0 & 50.8 & 71.4 & 48.8\\
        EMCEI & 32.6 & 44.6 & 56.0 & 51.2 & 41.6 & 42.6 & 43.0 & 52.6 & 52.4 & 52.0 & 71.2 & 49.1\\
        \hdashline
        SoT~(\textit{Ours}) & \textbf{51.2} & 51.0 & \textbf{66.8} & 67.4 & 50.6 & 50.0 & 52.0 & 61.4 &\textbf{58.8} &\textbf{61.2} & 76.8 & 58.8 \\
        ~~+3-shot & 46.2 & \textbf{52.6} & 63.2 & \textbf{68.4} & 49.6 & \textbf{52.0} & 52.0 & 54.8 & 57.0 & 55.2 & 76.8 & 57.1 \\
        ~~+CoT & 49.8 & 51.4 & 65.4 & 68.2 & \textbf{53.4} & 49.8 & \textbf{54.6} &\textbf{62.2}& 56.4 & \textbf{61.2} & \textbf{78.4} & \textbf{59.2} \\
        \midrule
        (\textbf{\texttt{post-training}}) \\
        xCoT & 44.6 & 32.4 & 55.8 & 57.0 & 21.2 & 31.4 & 29.8 & 59.2 & 33.8 & 54.2 & 65.4 & 44.1 \\
        QAlign & 43.0 & 31.2 & 53.4 & 53.4 & 22.2 & 28.4 & 33.8 & 59.8 & 21.4 & 49.6 & 71.0 & 42.5 \\
        MindMerger & 41.8 & 32.6 & 53.2 & 56.8 & 21.4 & 32.0 & 32.4 & 51.4 & 33.6 & 54.8 & 65.8 & 43.3 \\
        MAPO & 41.2 & 35.0 & 51.4 & 54.8 & 22.0 & 29.8 & 35.0 & 52.0 & 35.0 & 51.2 & 61.2 & 42.6 \\
    \midrule
    \midrule
    (\textbf{\texttt{training-free}}) & \multicolumn{12}{c}{(\texttt{Qwen2.5-7B-Instruct})}\\
        Direct & 8.6 & 15.0 & 5.0 & 9.8 & 17.0 & 4.4 & 14.8 & 8.4 & 3.6 & 8.0 & 8.8 & 9.4 \\
        DoLa & 53.8 & 44.6 & 75.4 & 65.8 & 30.6 & 41.8 & 52.4 & 63.0 & 74.0 & 74.6 & 75.4 & 59.2 \\
        SL-D & 61.4 & 52.4 & 73.8 & 74.4 & 35.0 & 43.4 & 52.2 & 63.8 & 73.4 & 72.4 & 74.4 & 61.5 \\
        DIP & 61.4 & 59.0 & 81.2 & 83.0 & 46.8 & 47.2 & 58.6 & 73.2 & 76.8 & 74.4 & 74.2 & 66.9 \\
        CLP & 64.2 & 54.8 & 75.6 & 70.0 & 41.0 & 41.6 & 53.0 & 64.6 & 65.2 & 75.6 & 70.2 & 61.4 \\
        EMCEI & 61.6 & 51.8 & 72.6 & 74.2 & 44.8 & 45.2 & 50.2 & 70.8 & 71.4 & 71.8 & 72.8 & 62.5 \\
        \hdashline
        SoT~(\textit{Ours}) & \textbf{65.0} & 58.2 & 82.6 & 83.8 & \textbf{49.6 }& 50.8 & 60.8 & 73.4 & 78.6 & 83.2 & 81.0 & 69.7 \\
        ~~+3-shot & 63.2 & 59.0 & \textbf{83.4} & 84.2 & \textbf{49.6} & 50.4 & 58.4 & \textbf{78.8} & \textbf{80.8} & 80.8 & \textbf{86.0} & \textbf{70.4} \\
        ~~+CoT & 64.0 & \textbf{59.6} & 81.2 & \textbf{86.4} & 47.2 & \textbf{51.8} & \textbf{61.2} & 75.4 & 79.4 & \textbf{83.8} & 83.6 & 70.3  \\
        \midrule
        (\textbf{\texttt{post-training}}) \\
        xCoT & 47.2 & 48.0 & 65.2 & 68.4 & 21.0 & 45.2 & 32.2 & 69.2 & 33.6 & 64.4 & 65.0 & 50.9 \\
        QAlign & 45.8 & 42.0 & 62.2 & 60.2 & 22.8 & 35.2 & 45.6 & 69.6 & 21.4 & 58.2 & 73.4 & 48.8 \\
        MindMerger & 44.2 & 33.2 & 57.2 & 67.0 & 21.2 & 42.4 & 38.6 & 57.6 & 34.4 & 64.6 & 65.0 & 47.8 \\
        MAPO & 40.2 & 38.0 & 61.2 & 62.0 & 22.4 & 42.8 & 43.4 & 58.4 & 35.8 & 62.0 & 60.8 & 47.9 \\
        
        \midrule
        \midrule
        (\textbf{\texttt{training-free}}) & \multicolumn{12}{c}{(\texttt{gpt-3.5-turbo})}\\
        Direct & 48.2 & 49.6 & 33.8 & 36.8 & 50.2 & 47.0 & 37.8 & 46.0 & 43.4 & 44.8 & 37.0 & 43.1 \\
        DoLa & 73.2 & 53.6 & 74.6 & 78.6 & 39.0 & 52.4 & 50.4 & 64.6 & 72.0 & 72.2 & 78.8 & 64.5 \\
        DIP & 75.4 & 60.8 & 81.2 & 81.4 & 43.6 & 54.8 & 62.6 & 70.0 & 79.2 & 73.8 & 77.8 & 69.1 \\
        CLP & 70.2 & 58.0 & 73.2 & 81.2 & 40.0 & 56.0 & 52.6 & 64.0 & 73.2 & 67.0 & 72.2 & 64.3 \\
        EMCEI & 78.6 & 62.0 & 83.0 & 84.6 & 44.2 & 57.6 & 62.4 & 72.8 & 82.8 & 75.4 & 79.8 & 71.2 \\
        \hdashline
        SoT~(\textit{Ours})  & 82.0 & \textbf{66.4} & 84.0 & \textbf{88.2} & 49.0 & 74.4 & 58.2 & 77.0 & \textbf{84.2} & 81.4 & 84.6 & 75.4 \\
        ~~+3-shot & 79.2 & 60.4 & 76.2 & 67.2 & \textbf{55.0} & 73.6 & \textbf{63.8} & 69.4 & 72.8 & 64.8 & 78.2 & 69.1 \\
        ~~+CoT  & \textbf{83.4} & 66.2 & \textbf{86.0} & 87.6 & 54.0 & \textbf{76.0} & 63.4 & \textbf{78.4} & 82.8 & \textbf{82.0} & \textbf{87.6} & \textbf{77.0} \\
        \midrule
        \midrule
        (\textbf{\texttt{training-free}}) & \multicolumn{12}{c}{(\texttt{Qwen2.5-32B-Instruct})}\\
        Direct & 19.2 & 27.4 & 16.6 & 18.6 & 23.2 & 23.2 & 21.2 & 19.4 & 9.4 & 25.0 & 5.4 & 19.0 \\
        DoLa & 78.4 & 66.2 & 75.4 & 84.6 & 40.4 & 54.8 & 67.0 & 82.8 & 84.8 & 81.6 & 83.4 & 72.7 \\
        SL-D & 61.4 & 52.4 & 73.8 & 74.4 & 35.0 & 43.4 & 52.2 & 63.8 & 73.4 & 72.4 & 74.4 & 61.5 \\
        DIP & 71.2 & 68.1 & 83.2 & 89.4 & 49.4 & 55.6 & 68.2 & 87.0 & 89.6 & 88.2 & 89.6 & 76.3 \\
        CLP & 75.4 & 70.2 & 81.2 & 92.8 & 43.8 & 55.4 & 70.8 & 81.8 & 85.2 & 83.8 & 84.4 & 75.0 \\
        EMCEI & 78.6 & 62.0 & 83.0 & 84.6 & 44.4 & 57.6 & 62.4 & 82.8 & 82.8 & 91.2 & 85.8 & 74.1 \\
        \hdashline
        SoT~(\textit{Ours}) & \textbf{87.0} & 74.8 & 96.2 & 97.2 & \textbf{53.8} & 65.4 & 78.4 & 91.0 & 95.0 & 95.8 & \textbf{96.0} & 84.6 \\
        ~~+3-shot MindMerger & 86.6 & \textbf{78.2} & \textbf{96.8} & \textbf{97.8} & 52.2 & \textbf{69.8} & \textbf{79.8} & 88.6 & \textbf{95.6} & \textbf{96.4} & 95.4 & \textbf{85.2} \\
        ~~+CoT & 86.4 & 76.0 & 96.6 & 96.8 & 53.6 & 67.2 & 78.8 & \textbf{91.6} & 95.4 & 95.8 & \textbf{96.0} & 84.9 \\
        \midrule
        (\textbf{\texttt{post-training}}) \\
        xCoT & 67.6 & 68.2 & 84.2 & 86.4 & 24.2 & 55.2 & 52.2 & 73.4 & 73.4 & 77.6 & 80.2 & 67.5 \\
        QAlign & 65.2 & 62.6 & 82.0 & 80.0 & 24.8 & 55.2 & 55.6 & 77.6 & 52.8 & 80.0 & 84.2 & 65.5 \\
        MindMerger & 70.6 & 51.8 & 81.2 & 83.4 & 25.8 & 51.0 & 42.4 & 73.4 & 54.4 & 75.2 & 80.4 & 62.7 \\
        MAPO & 67.2 & 68.8 & 82.6 & 87.2 & 24.0 & 59.0 & 57.2 & 74.8 & 59.2 & 72.6 & 81.0 & 66.7 \\
        
        \bottomrule
    \end{tabular}%
     \caption{Supplementary results~(\%) of commonsense reasoning on XCOPA using various LLMs, including \textbf{\texttt{DeepSeek-R1-7B}}, \textbf{\texttt{Qwen2.5-32B-Instruct}}, \textbf{\texttt{gpt-3.5-turbo}} and \textbf{\texttt{Qwen2.5-32B-Instruct}}.}
    \label{tab:supresults_xcopa}
\end{table*}

\end{CJK*}
\end{document}